# Dynamic interactive group decision making method on two-dimensional language evaluation information

**Yukun Zhang**

The Chinese University Of  HongKong    215010026@link.cuhk.edu.cn

**Abstract.** The language evaluation information of the interactive group decision method at present is based on the one-dimension language variable. At the same time, multi-attribute group decision making method based on two-dimension linguistic information only use single-stage and    static evaluation method. In this paper, we propose a dynamic group decision making method based on two-dimension linguistic information, combining dynamic interactive group decision making methods with two-dimensional language evaluation information The method first use Two-Dimensional Uncertain Linguistic Generalized Weighted Aggregation (DULGWA) Operators to aggregate the preference information of each decision maker, then adopting  dynamic information entropy method to obtain  weights of attributes at each stage. Finally we propose the group consistency index to quantify the termination conditions of group interaction. One example is given to verify the developed approach and to demonstrate its effectiveness.

# 1.Introduction

Group decision-making methods have wide applications in many fields, such as economic, science, culture and other aspects .As for group decision-making problems ,decision maker have difficulties expressing the attribute values by the real numbers owing to the complexity and uncertainty of human thinking .while it may be easier to express the attribute values by linguistic information. These problems are called Uncertain Multiple Attribute Decision Making (UMADM) problems. The research of Multiple Attribute Decision Making(MADM) problems theory have made great progress up to now.As for UMADM theory, Zedeh [1] firstly proposed the definition of the linguistic variables. Zhu et al.[3] introduced the 2-dimension linguistic evaluation information to solve the fuzzy problems .which means I class linguistic information and II class linguistic information are used to evaluate the objects. I class linguistic information describe the evaluation results of attributes given by decision makers,while II class linguistic information gives the subjective evaluation of decision makers on the reliability of there evaluation,hence we call the 2-dimension linguistic assessment information as 2-dimension linguistic variables, which are more Accurate to express the opinion of decision makers for attribute values.

Up to now,there are some studies of 2-dimension linguistic information method, Zhu et al.[3] introduced the definition of 2-dimension linguistic information and apply the framework into subjective evidencial reasoning in the area of group decision making, but this method does not deal with the multi-criteria decision problems. Liu and Zhang [4] further extend 2-dimension linguistic variables to 2-dimension uncertain linguistic variables in which the attribute weight is unknown and 2-dimension uncertain linguistic information is uncertain, and propose a new method to deal with multiple attribute group decision making problems. Liu and Yu [5] proposed 2-dimension uncertain linguistic power generalized aggregation operators(2DULPGA) and 2-dimension uncertain linguistic power generalized weighted aggregation operators(2DULPGWA) combining the power average(PA) operator with 2-dimension linguistic information, and discuss some properties and special cases of them. Then they proposed group decision making method based on some power generalized aggregation operators with 2-dimension uncertain linguistic information.

There is no doubt that the research on multiple attribute of group decision making problems based on 2-dimension linguistic information have great significance, but in the real decision making problems,it does not consider the dynamic and interactive case. At present, the main research achievements based on dynamic group decision making method have ,made the fruitful achievements. For example,Zhang [5] discusses a method of group evaluation based on multi-phase dynamic interactive and then defines stability and consistency of indicator indexes to explore the interaction.

Obviously , there are two shortcomings of the group decision-making method.Firstly, the main research of traditional dynamic group decision making method is focus on the one dimension and the number value and it is difficult for one dimension information to express the fuzzy assessment . And the multi-attribute decision making on the 2-dimension uncertain linguistic information just at present stop at the single stage and do not take the dynamic decision-making perspective into consideration. In view of the above two shortcomings ,we combine the 2-dimension linguistic variables with dynamic group evaluation and propose the multi-round dynamic group decision making method on 2-dismension uncertain linguistic variables and apply them to solve the UMSDM problems.

The reminder of this paper is shown as follows. In section 1, we give an brief introduction of the research background. In Section 2, we introduced some basic concepts of 2-dimension uncertain linguistic information. In Section 3, we brief review some power aggregation operators based on 2-dimension uncertain linguistic variables. In Section 4, we introduce the basic principles of dynamic interactive group decision making method based on 2-dimension uncertain linguistic variables .In Section 5, we give the decision making steps of dynamic interactive group decision. making method. In Section 6, we give one example to illustrate the decision making steps based on the proposed method .In the last Section , we give the concluding remarks and future research directions.

## 2. Preliminaries: $[s_{a1},s_{b1}] \oplus [s_{a2},s_{b2}] = [s_{a1+a2},s_{b1+b2}]$

### 2.1 uncertain linguistic variable

When decision makers give the evaluation of the objects, generally it is essential to set an appropriate linguistic assessment in advance. Suppose $\bar{s}=[s_a,s_b]$, $s_a,s_b \in \bar{S}$ 且 $a \leq b$, $s_a, s_b$ are the lower limit and upper limit of $\bar{s}$ respectively, then we call $\bar{s}$ an uncertain linguistic variable. Suppose we use $\bar{S}$ to express the set of all uncertain linguistic variables.

**Definition 1,** Let $\bar{s}_1 = [s_{a1},s_{b1}]$, $\bar{s}_2 = [s_{a2},s_{b2}]$ be any two uncertain linguistic variables, then the operational rules are shown as follows:

（1） $\bar{s}_1 \oplus \bar{s}_2 =$  (1)

（2） $\bar{s}_1 \otimes \bar{s}_2 = [s_{a1},s_{b1}] \otimes [s_{a2},s_{b2}] = [s_{a1 \times a2},s_{b1 \times b2}]$  (2)

（3） $\bar{s}_1 / \bar{s}_2 = [s_{a1},s_{b1}]/[s_{a2},s_{b2}] = [s_{a1/b2},s_{b1/a2}]$  (3)

（4） $\lambda \bar{s}_1 = \lambda [s_{a1},s_{b1}] = [s_{\lambda*a1},s_{\lambda*b1}]$  (4)

（5） $\lambda(\bar{s}_1 \oplus \bar{s}_2) = \lambda \bar{s}_1 \oplus \lambda \bar{s}_2$  (5)

（6） $(\lambda_1 + \lambda_2)\bar{s}_1 = \lambda_1 \bar{s}_1 \oplus \lambda_2 \bar{s}_2$  (6)

### 2.2 The definition of two-dimension uncertain linguistic variable

**Definition** 2: let $\hat{s} = ([\dot{s}_a,\dot{s}_b],[\ddot{s}_c,\ddot{s}_d])$, where $[\dot{s}_a,\dot{s}_b]$ is I class uncertain linguistic information, which represents decision maker's judgement to evaluated object, and $[\dot{s}_a,\dot{s}_b]$ are the elements from the predefined linguistic assessment set $S_I = (\dot{s}_0,\dot{s}_1,\cdots,\dot{s}_{l-1})$ while $[\ddot{s}_c,\ddot{s}_d]$ is II class uncertain linguistic information, which represents the subjective evaluation on the reliability of their given results. And $[\ddot{s}_c, \ddot{s}_d]$ are the elements from the predefined linguistic assessments set $S_{II} = (\ddot{s}_0, \ddot{s}_1, \cdots, \ddot{s}_{z-1})$, then $\hat{s}$ is called the 2-dimension uncertain linguistic variable。

## 2.3. The operational rules and the characteristic of two dimension uncertain linguistic variables

For any two 2-dimension uncertain linguistic variables $\hat{s}_1 = ([\dot{s}_{a1}, \dot{s}_{b1}], [\ddot{s}_{c1}, \ddot{s}_{d1}])$ and $\hat{s}_2 = ([\dot{s}_{a2}, \dot{s}_{b2}], [\ddot{s}_{c2}, \ddot{s}_{d2}])$, the operational rules are given shown as follows:

(1) $\hat{s}_1 \oplus \hat{s}_2 = ([\dot{s}_{a1}, \dot{s}_{b1}], [\ddot{s}_{c1}, \ddot{s}_{d1}]) \oplus ([\dot{s}_{a2}, \dot{s}_{b2}], [\ddot{s}_{c2}, \ddot{s}_{d2}])$

$$= ([\dot{s}_{a1+a2}, \dot{s}_{b1+b2}], [\ddot{s}_{\min(c1,c2)}, \ddot{s}_{\min(d1,d2)}]) \quad (7)$$

(2) $\hat{s}_1 \otimes \hat{s}_2 = ([\dot{s}_{a1}, \dot{s}_{b1}], [\ddot{s}_{c1}, \ddot{s}_{d1}]) \otimes ([\dot{s}_{a2}, \dot{s}_{b2}], [\ddot{s}_{c2}, \ddot{s}_{d2}])$

$$= ([\dot{s}_{a1 \times a2}, \dot{s}_{b1 \times b2}], [\ddot{s}_{\min(c1,c2)}, \ddot{s}_{\min(d1,d2)}]) \quad (8)$$

(3) $\hat{s}_1 / \hat{s}_2 = ([\dot{s}_{a1}, \dot{s}_{b1}], [\ddot{s}_{c1}, \ddot{s}_{d1}]) / ([\dot{s}_{a2}, \dot{s}_{b2}], [\ddot{s}_{c2}, \ddot{s}_{d2}])$

$$= ([\dot{s}_{a1/b2}, \dot{s}_{b1/a2}], [\ddot{s}_{\min(c1,c2)}, \ddot{s}_{\min(d1,d2)}]), a_2, b_2 \neq 0 \quad (9)$$

(4) $\lambda \hat{s}_1 = \lambda([\dot{s}_{a1}, \dot{s}_{b1}], [\ddot{s}_{c1}, \ddot{s}_{d1}]) = \lambda([\dot{s}_{\lambda \times a1}, \dot{s}_{\lambda \times b1}], [\ddot{s}_{c1}, \ddot{s}_{d1}])(\lambda \geq 0)$ (10)

(5) $(\hat{s}_1^\lambda) = ([\dot{s}_{a1}, \dot{s}_{b1}], [\ddot{s}_{c1}, \ddot{s}_{d1}])^\lambda = \lambda([\dot{s}_{(a1)^\lambda}, \dot{s}_{(b1)^\lambda}], [\ddot{s}_{c1}, \ddot{s}_{d1}])(\lambda \geq 0)$ (11)

Let $\hat{s}_1 = ([\dot{s}_{a1}, \dot{s}_{b1}], [\ddot{s}_{c1}, \ddot{s}_{d1}])$, $\hat{s}_2 = ([\dot{s}_{a2}, \dot{s}_{b2}], [\ddot{s}_{c2}, \ddot{s}_{d2}])$ and $\hat{s}_3 = ([\dot{s}_{a3}, \dot{s}_{b3}], [\ddot{s}_{c3}, \ddot{s}_{d3}])$ be any three 2-dimension uncertain linguistic variables and $\lambda_1, \lambda_2, \lambda_3 \geq 0$.

The two-dimension uncertain linguistic variables satisfy the following properties.

(1) $\hat{s}_1 \oplus \hat{s}_2 = \hat{s}_2 \oplus \hat{s}_1$ (12)

(2) $\hat{s}_1 \otimes \hat{s}_2 = \hat{s}_2 \otimes \hat{s}_1$ (13)

(3) $\hat{s}_1 \oplus \hat{s}_2 \oplus \hat{s}_3 = \hat{s}_1 \oplus (\hat{s}_2 \oplus \hat{s}_3)$ (14)

(4) $\hat{s}_1 \otimes \hat{s}_2 \otimes \hat{s}_3 = \hat{s}_1 \otimes (\hat{s}_2 \otimes \hat{s}_3)$ (15)

(5) $\hat{s}_1 \otimes (\hat{s}_2 \oplus \hat{s}_3) = (\hat{s}_1 \otimes \hat{s}_2) \oplus (\hat{s}_1 \otimes \hat{s}_3)$ (16)

(6) $\lambda(\hat{s}_1 \oplus \hat{s}_2) = (\lambda\hat{s}_1) \oplus (\lambda\hat{s}_2)$ （17）

(7) $(\lambda_1 + \lambda_2)\hat{s}_1 = (\lambda\hat{s}_1) \oplus (\lambda\hat{s}_2)$ （18）

**2.4. The expectation of the two-dimension uncertain linguistic variable**

**Definition 3** Let $\hat{s}_1 = ([\dot{s}_{a1}, \dot{s}_{b1}], [\ddot{s}_{c1}, \ddot{s}_{d1}])$ be a two-dimension uncertain linguistic variable, $[\dot{s}_a, \dot{s}_b]$ are the elements from the predefined linguistic assessment set $S_I = (\dot{s}_0, \dot{s}_1, \cdots, \dot{s}_{l-1})$ and $[\ddot{s}_c, \ddot{s}_d]$ are the elements from the predefined linguistic assessment set $S_{II} = (\ddot{s}_0, \ddot{s}_1, \cdots, \ddot{s}_{t-1})$, then the expectation $E(\hat{s}_1)$ of $\hat{s}_1$ is defined as

$$E(\hat{s}_1) = \frac{a_1 + b_1}{2*(l-1)} * \frac{c_1 + d_1}{2*(z-1)} \quad (19)$$

**2.5. The comparison of two-dimension uncertain linguistic variables**

Let $\hat{s}_1 = ([\dot{s}_{a1}, \dot{s}_{b1}], [\ddot{s}_{c1}, \ddot{s}_{d1}])$ and $\hat{s}_2 = ([\dot{s}_{a2}, \dot{s}_{b2}], [\ddot{s}_{c2}, \ddot{s}_{d2}])$ be any two-dimension uncertain linguistic variables, if $E(\hat{s}_1) \geq E(\hat{s}_2)$, then $\hat{s}_1 \geq \hat{s}_2$, or vice versa

**2.6. The distance between two-dimension uncertain linguistic variables**

Let $\hat{s}_1 = ([\dot{s}_{a1}, \dot{s}_{b1}], [\ddot{s}_{c1}, \ddot{s}_{d1}])$, $\hat{s}_2 = ([\dot{s}_{a2}, \dot{s}_{b2}], [\ddot{s}_{c2}, \ddot{s}_{d2}])$ be any two-dimension uncertain linguistic variables, and $f: \hat{S} \times \hat{S} \to R$。If $d(\hat{s}_1, \hat{s}_2)$ satisfies the following condition

(1) $0 \leq d(\hat{s}_1, \hat{s}_2) \leq 1, d(\hat{s}_1, \hat{s}_2) = 0$ （20）

(2) $d(\hat{s}_1, \hat{s}_2) = d(\hat{s}_2, \hat{s}_1)$ （21）

(3) $d(\hat{s}_1, \hat{s}_2) + d(\hat{s}_2, \hat{s}_3) \geq d(\hat{s}_1, \hat{s}_3)$ （22）

then $d(\hat{s}_1, \hat{s}_2)$ is the distance between two-dimension uncertain linguistic variables, the Hamming distance of $\hat{s}_1, \hat{s}_2$ is defined as follows:

$$d(\hat{s}_1, \hat{s}_2) = \frac{1}{4(l-1)} \left( \left| a_1 \times \frac{c_1}{z-1} - a_2 \times \frac{c_2}{z-1} \right| + \left| a_1 \times \frac{d_1}{z-1} - a_2 \times \frac{d_2}{z-1} \right| \right.$$

$$\left. + \left| b_1 \times \frac{c_1}{z-1} - b_2 \times \frac{c_2}{z-1} \right| + \left| b_1 \times \frac{d_1}{z-1} - b_2 \times \frac{d_2}{z-1} \right| \right) \quad (23)$$

**3. The two-dimension uncertain linguistic generalized aggregation operators**

Definition Let $\hat{s}_k = ([\dot{s}_{ak}, \dot{s}_{bk}], [\ddot{s}_{ck}, \ddot{s}_{dk}])$ be the collection of two-dimension

uncertain linguistic variables, and $2DULGWA:\Omega^p \to \Omega$. If

$$2DULGWA(\hat{s}_1,\hat{s}_2,\ldots,\hat{s}_p)=\left(\sum_{k=1}^{p}w_k\hat{s}_k^{\alpha}\right)^{1/\alpha} \quad (24)$$

where $\Omega$ is the set of all two-dimension uncertain li9nguistic variables, and $w=(w_1,w_2,\ldots,w_p)^T$ is the weight vector of $\hat{s}_k(k=1,2,\ldots p)$, satisfying $w_k \geq 0$, $\sum_{k=1}^{p}w_k=1$. $\alpha$ is a parameter such that $\alpha \in (-\infty,0)\cup(0,+\infty)$, then 2DULGWA is called the 2-dimensional uncertain linguistic generalized aggregation(2DULGWA) operator based on the operational rules of the 2-dimension uncertain linguistic variables, we have the following theorems.

**Theorem 1** let $\hat{s}_k = ([\dot{s}_{ak},\dot{s}_{bk}],[\ddot{s}_{ck},\ddot{s}_{dk}])$ be the collection of two-dimension uncertain linguistic variables, then

$$2DULGWA(\hat{s}_1,\hat{s}_2,\ldots,\hat{s}_p)=\left(\sum_{k=1}^{p}w_k\hat{s}_k^{\alpha}\right)^{1/\alpha}=\left(\left[\dot{s}_{\left(\sum_{k=1}^{p}w_k a_k^{\alpha}\right)^{1/\alpha}},\dot{s}_{\left(\sum_{k=1}^{p}w_k b_k^{\alpha}\right)^{1/\alpha}}\right],\left[\ddot{s}_{\min_k c_k},\ddot{s}_{\min_k d_k}\right]\right)$$

(25）

**Theorem2  idempotency :**
Let $\hat{s}_k = \hat{s}, (\hat{s}_k = ([\dot{s}_a,\dot{s}_b],[\ddot{s}_c,\ddot{s}_d]))$, then

$$2DULGWA(\hat{s}_1,\hat{s}_2,\ldots,\hat{s}_p)= \hat{s}$$

**Theorem3 Commutativity :** let $(\hat{s}_1',\hat{s}_2',\ldots,\hat{s}_p')$ be any permutation of $(\hat{s}_1,\hat{s}_2,\ldots,\hat{s}_p)$, then

$$2DULGWA(\hat{s}_1,\hat{s}_2,\ldots,\hat{s}_p)= 2DULGWA(\hat{s}_1',\hat{s}_2',\ldots,\hat{s}_p')$$

**Thoerem4  Monotomicity**：if $\hat{s}_k \leq \hat{s}_k'$ for all k，then

$$2DULGWA(\hat{s}_1,\hat{s}_2,\ldots,\hat{s}_p)\leq 2DULGWA(\hat{s}_1',\hat{s}_2',\ldots,\hat{s}_p')$$

**Theorem5  (Boundedness)**
The 2DULGWA operator lies between the max and min operator, i.e.,

$$\min(\hat{s}_1,\hat{s}_2,\ldots,\hat{s}_p)\leq 2DULGWA(\hat{s}_1,\hat{s}_2,\ldots,\hat{s}_p)\leq \max(\hat{s}_1,\hat{s}_2,\ldots,\hat{s}_p)$$

$$2DULGWA(\hat{s}_1,\hat{s}_2,\ldots,\hat{s}_p)= \sum_{k=1}^{p}w_k\hat{s}_k$$

When $\alpha = 1$, 2DUKGWA operator can be reduced to the 2-dimension uncertain linguistic weighted aggregation (2DULWA) operator.

## 4 An approach to dynamic interactive group decision making based on two-dimensional language evaluation information

### 4.1 The Description of the Dynamic Interactive Group Decision Making Problem

Suppose a multiple attribute dynamic interactive group decision making problem with 2-dimension uncertain linguistic information. Let $A = \{A_1, A_2, \cdots, A_m\}$ be the set of alternatives, $C = \{C_1, C_2, \cdots C_n\}$ be the set of attributes, and the attributes weights vector will be $w_1 = (w_1, w_2 \cdots w_n)$, satisfying $0 \leq w_j \leq 1, \sum_{j=1}^{n} w_j = 1$, where $w_j$ donate the important degree of attributes $C_j$. Let $E = \{e_1, e_2, \cdots e_p\}$ be the set of decision makers, and $\lambda = (\lambda_1, \lambda_2 \cdots \lambda_p)$ is the expert weight vector, satisfying $0 \leq \lambda_k \leq 1, \sum_{k=1}^{p} \lambda_k = 1$. Suppose that $s_{ij}^{kt} = ([x_{ij}^{hk}, x_{ij}^{uk}], [g_{ij}^{hk}, g_{ij}^{uk}])$ is the attribute value in the attribute $C_j$ with respect to the alternative $A_i$ where $([x_{ij}^{hk}, x_{ij}^{uk}])$ is the I class of uncertain linguistic variables, and $([g_{ij}^{hk}, g_{ij}^{uk}])$ is the II class of uncertain linguistic variables. Suppose after the $l(l \geq 3)$ round interactions, $s_{ij}^{kt} = ([x_{ij}^{hk}, x_{ij}^{uk}], [g_{ij}^{hk}, g_{ij}^{uk}])$ $(i = 1, 2, \cdots, m; j = 1, 2, \cdots, n; k = 1, 2, \cdots, p; d = 1, 2, \cdots, l)$ is given by the the decision maker $e_k^t$ in the $t$ round for alternative $A_i$ with respect to attribute $C_j$, , and $\hat{S}_k^t = [s_{ij}^{kt}]_{m \times n}$ is the decision matrix given by the decision maker, i.e.

$$\widehat{S}_k^t = \begin{bmatrix} \hat{s}_{11}^{kt} & \hat{s}_{11}^{kt} & \cdots & \hat{s}_{1n}^{kt} \\ \hat{s}_{21}^{kt} & \hat{s}_{21}^{kt} & \cdots & \hat{s}_{2n}^{kt} \\ \vdots & \vdots & \ddots & \vdots \\ \hat{s}_{m1}^{kt} & \hat{s}_{m1}^{kt} & \cdots & \hat{s}_{mn}^{kt} \end{bmatrix}$$

the goal of this decision making method is to rank the alternatives

### 4.2 The Principle of Group Consistency Index

Suppose let $E = \{e_1, e_2, \cdots e_p\}$ be the set of decision maker, Let $A = \{A_1, A_2, \cdots, A_m\}$ be the set of alternatives, $C = \{C_1, C_2, \cdots C_n\}$ be the set of attributes，The individual preference matrix of decision maker $e_k^t$ is $\hat{X}_k^t = \left[\hat{s}_{ij}^{kt}\right]_{m \times n}$, such that $X_k^t = \begin{bmatrix} \hat{s}_{11}^{kt} & \hat{s}_{11}^{kt} & \cdots & \hat{s}_{1n}^{kt} \\ \hat{s}_{21}^{kt} & \hat{s}_{21}^{kt} & \cdots & \hat{s}_{2n}^{kt} \\ \vdots & \vdots & \ddots & \vdots \\ \hat{s}_{m1}^{kt} & \hat{s}_{m1}^{kt} & \cdots & \hat{s}_{mn}^{kt} \end{bmatrix}$. The group decision preference matrix is $\hat{Y}^t = \left[\hat{y}_{ij}^t\right]_{m \times n}$, such that $\hat{Y}^t = \begin{bmatrix} \hat{y}_{11}^t & \hat{y}_{12}^t & \cdots & \hat{y}_{1n}^t \\ \hat{y}_{21}^t & \hat{y}_{22}^t & \cdots & \hat{y}_{2n}^t \\ \vdots & \vdots & \ddots & \vdots \\ \hat{y}_{m1}^t & \hat{y}_{m2}^t & \cdots & \hat{y}_{mn}^t \end{bmatrix}$, which is affected by the group decision rule $h$.

Definition: $\eta_k^t$ is defined as follows:

$$\eta_k^t = \sum_{i=1}^{m}\sum_{j=1}^{n} d\left(\hat{s}_{ij}^{kt}, \hat{y}_{ij}^{kt}\right) / E\left(\hat{s}_{ij}^{kt}\right) \qquad (26)$$

then $\eta_k^t$ is called group consistency indicator between the individual preference matrix of decision maker $e_k^t$ and group decision preference matrix. $\eta_k^t$ reflect degree of consistency and difference between individual decision matrix and group decision matrix. The smaller the $\eta_k^t$ is, the smaller the difference between individual decision matrix and group decision matrix is. Different decision makers' consistency indicators $\eta_k^t$ constitute a vector $\eta_t = \left(\eta_1^t, \eta_2^t, \cdots \eta_p^t\right)$, The aim of dynamic interactive group decision making is to require the consistency indicators of each decision maker to reach a consistent condition, which is defines as:

$\eta_k^t \leq \zeta_k$,

where $\zeta_k$ is the threshold of the group consistency indicator set for each decision maker。

When using the group consistency index, the judgment of the consistency between the individual preference information and the group preference decision information can be quantified, which is beneficial to the smooth progress of the group interaction process. After T round interactions, the Value of group consistency index vector $\eta_t = \left(\eta_1^t, \eta_2^t, \cdots \eta_p^t\right)$ should become gradually decrease, which means the group interaction has gradually improved the consistency of each decision maker, and such a group interaction process is effective.

### 4.3. Objective Weights Of Attributes Based On Entropy

The more consistent the expert group evaluates the same scheme under the same attribute, the more effective the group decision on the attribute is, the more weight should be assigned to the attribute, and the more invalid, the smaller the weight should be assigned to the attribute. The following information entropy is used to measure the consistency of the evaluation results of the expert group under a certain attribute. The larger the entropy value is, the larger the attribute weight is. The entropy value can be normalized to obtain the objective weight of the attribute.

The specific calculation method is as follows:

Step 1.calculate the expectation of two-dimension uncertain linguistic variables

Step 2.calculate the information entropy of the decision makers in the t round for alternative $A_i$ with respect to attribute $C_j$.

$$I_{ij}^t = -\frac{1}{\ln p}\sum_{k=1}^{p} \frac{E(s_{ij}^{kt})}{\sum_{h=1}^{p} E(s_{ij}^{kt})} \ln \frac{E(s_{ij}^{kt})}{\sum_{h=1}^{p} E(s_{ij}^{kt})}$$ （27）

Step3: calculate the average value of information entropy with respect to attribute $C_j$

$$I_j^t = \frac{1}{m}\sum_{i=1}^{m} I_{ij}^t$$ （28）

step4: calculate the weights of attributes based on entropy

$$w_j^t = \frac{I_j^t}{\sum_{j=1}^{n} I_j^t}$$ （29）

### 4.4. Adjustment of decision makers' preference information：

In order to avoid the randomness of expert preference adjustment, we regulates its preference adjustment method.

Definition:

$$\hat{s}_{ij}^{k(t+1)} = f(\hat{s}_{ij}^{kt}, \hat{y}_{ij}^{kt})$$ （30）

$$\hat{S}_k^{t+1} = f(\hat{S}_k^t, \hat{Y}_k^t)$$ （31）

The equation reflect group interaction and adjustment of the individual preference information of the T+1 round group interaction according to the preference evaluation information $\hat{S}_k^t, \hat{Y}_k^t$ of the t-th round group interaction

Suppose $f(\hat{s}_{ij}^{kt}, \hat{y}_{ij}^{kt})$ is the linear function, which is defined as follows:

$$\hat{s}_{ij}^{k(t+1)} = (1-w_{ij}^t)\hat{s}_{ij}^{kt} + w_{ij}^t \hat{y}_{ij}^t$$ （32）

where $w_{ij}^t$ reflect the degree of emphasis or concession on group preference information, and $(1-w_{ij}^t)$ reflects the degree of adherence of decision makers to their opinions.

In this way, each decision maker gives the initial preference information.and then in each round of the group interaction process, the value of individual's preference information evaluation is adjusted according to adjusted function , and then the group consistency is achieved.

## 4.5 The method of dynamic interactive group decision making based on two-dimensional language evaluation information

In this section, we will use the dynamic interactive group decision making method based on two-dimensional language evaluation information to solve the multiple attribute decision making problem. The steps of the proposed method will be introduced as follows:

**Step 1:** calculate the comprehensive value of each alternative.
Based on the 2DULGA, we can calculate the comprehensive value of each alternative, obtain the group decision preference matrix $\hat{Y}^t = [\hat{y}^t_{ij}]_{m \times n}$, where $\hat{y}^t_{ij} = ([x^{Ht}_{ij}, x^{Ut}_{ij}], [g^{Ht}_{ij}, g^{Ut}_{ij}])$

$$\hat{y}^t_{ij} = 2DULGWA(\hat{s}_1, \hat{s}_2, \ldots, \hat{s}_p) = \left(\sum_{k=1}^{p} w_k \hat{s}^\alpha_k\right)^{1/\alpha} = \left(\left[\dot{s}_{\left(\sum_{k=1}^{p} w_k a^\alpha_k\right)^{1/\alpha}}, \dot{s}_{\left(\sum_{k=1}^{p} w_k b^\alpha_k\right)^{1/\alpha}}\right], \left[\ddot{s}_{\min_k c_k}, \ddot{s}_{\min_k d_k}\right]\right)$$

（33）

**Step2:** calculate the distance matrix

According to equation (2-1), calculate the distance matrix $\hat{D}^t_k = [\hat{d}^{kt}_{ij}]_{m \times n}$ between the individual decision preference matrix $\hat{X}^t_k = [s^{kt}_{ij}]_{m \times n}$ and group decision preference matrix $\hat{Y}^t = [\hat{y}^t_{ij}]_{m \times n}$. $\hat{d}^{kt}_{ij}$ is Hamming distance between the two_dimension uncertain linguistic variables $s^{kt}_{ij}$ and $\hat{y}^t_{ij}$, which is defined by the formula where $D^t_k = \begin{bmatrix} \hat{d}^{kt}_{11} & \hat{d}^{kt}_{12} & \cdots & \hat{d}^{kt}_{1n} \\ \hat{d}^{kt}_{21} & \hat{d}^{kt}_{22} & \cdots & \hat{d}^{kt}_{2n} \\ \vdots & \vdots & \ddots & \vdots \\ \hat{d}^{kt}_{m1} & \hat{d}^{kt}_{m2} & \cdots & \hat{d}^{kt}_{mn} \end{bmatrix}$.

$$d(\hat{s}_1, \hat{s}_2) = \frac{1}{4(l-1)}\left(\left|a_1 \times \frac{c_1}{z-1} - a_2 \times \frac{c_2}{z-1}\right| + \left|a_1 \times \frac{d_1}{z-1} - a_2 \times \frac{d_2}{z-1}\right|\right.$$

$$\left. + \left|b_1 \times \frac{c_1}{z-1} - b_2 \times \frac{c_2}{z-1}\right| + \left|b_1 \times \frac{d_1}{z-1} - b_2 \times \frac{d_2}{z-1}\right|\right) \quad (34)$$

**Step 3:** calculate the expectation of two-dimensional language evaluation information $E(\hat{s}^{kt}_{ij})$

$$E\left(\hat{s}_{ij}^{kt}\right)=\frac{a_{ij}^{kt}+b_{ij}^{kt}}{2*(l-1)}*\frac{c_{ij}^{kt}+d_{ij}^{kt}}{2*(z-1)} \quad (35)$$

**Step 4：** calculate the group consistency index

According to the equation to obtain the vector of group consistency index $\eta_t=\left(\eta_1^t,\eta_2^t,\cdots\eta_p^t\right)$.

$$\eta_k^t=\sum_{i=1}^{m}\sum_{j=1}^{n}d\left(\hat{s}_{ij}^{kt},\hat{y}_{ij}^{kt}\right)/E\left(\hat{s}_{ij}^{kt}\right) \quad (36)$$

**Step 5:** Determine whether to perform group interaction

Determine whether the group consistency vector index satisfies the condition, which is $\eta_k^t\leq\zeta_k$,

If the above conditions are met, then the interaction is terminated. If not, calculate the information entropy to adjust preference matrix and then go to step 6.

**Step 6** :Calculate the entropy weight under each attribute according to the equation （4-1）,（4-2）,（4-3）

**Step7：** Adjust decision makers' preference information according to the equation （4-5）,（4-6）,（4-7）, then obtain ta new round of decision makers' individual decision preference information, return to step 1.

# 6.An illustrate example

In this section, we will provide one example. One is for evaluation for the supply chain and byt this example, we can explain the actual application of the proposed method.

This is an example of supply chain innovation ability evaluation. There are four supply chains $\{A_1,A_2,A_3,A_4\}$, which are evaluated by the following four attributes:flexible competitiveness of the supply chain$(C_1)$,brand competitiveness of the supply chain$(C_2)$,product competitiveness of the supply chain$(C_3)$,employee competitiveness of supply chain$(C_4)$.Three experts $\{e_1,e_2,e_3\}$ were invited to evaluate these supply chain by these four attributes.Suppose that $\lambda=(1/3,1/3,1/3)$ is the weight vector of the three experts. The attribute values given by the experts take the form of two-dimension uncertain linguistic variables and they are shown in table 4-1-1,4-1-2,4-1-3.The experts adopted I class linguistic set $S_I=(\dot{s}_0,\dot{s}_1,\dot{s}_2,\dot{s}_3,\dot{s}_4,\dot{s}_5,\dot{s}_6)$ and the II class linguistic set $S_{II}=(\ddot{s}_0,\ddot{s}_1,\ddot{s}_2,\ddot{s}_3,\ddot{s}_4)$. Termination condition of group interactions is $\xi_k=(1.5,2.5,1.5)$

**The First round of group decision**

Table 1: The index values with expects four supply chain by experts e1 in the 1st round

| Supply chain | Attribute $(c_1)$ | Attribute $(c_2)$ | Attribute $(c_3)$ | Attribute $(c_4)$ |
|---|---|---|---|---|
| $a_1$ | $([\dot{s}_5,\dot{s}_5],[\ddot{s}_2,\ddot{s}_3])$ | $([\dot{s}_2,\dot{s}_2],[\ddot{s}_3,\ddot{s}_3])$ | $([\dot{s}_4,\dot{s}_5],[\ddot{s}_4,\ddot{s}_4])$ | $([\dot{s}_3,\dot{s}_4],[\ddot{s}_1,\ddot{s}_2])$ |
| $a_2$ | $([\dot{s}_3,\dot{s}_4],[\ddot{s}_2,\ddot{s}_3])$ | $([\dot{s}_5,\dot{s}_5],[\ddot{s}_3,\ddot{s}_3])$ | $([\dot{s}_3,\dot{s}_3],[\ddot{s}_4,\ddot{s}_4])$ | $([\dot{s}_4,\dot{s}_4],[\ddot{s}_1,\ddot{s}_2])$ |
| $a_3$ | $([\dot{s}_2,\dot{s}_3],[\ddot{s}_2,\ddot{s}_3])$ | $([\dot{s}_3,\dot{s}_4],[\ddot{s}_3,\ddot{s}_3])$ | $([\dot{s}_3,\dot{s}_4],[\ddot{s}_4,\ddot{s}_4])$ | $([\dot{s}_4,\dot{s}_5],[\ddot{s}_1,\ddot{s}_2])$ |
| $a_4$ | $([\dot{s}_5,\dot{s}_6],[\ddot{s}_2,\ddot{s}_3])$ | $([\dot{s}_1,\dot{s}_2],[\ddot{s}_3,\ddot{s}_3])$ | $([\dot{s}_2,\dot{s}_3],[\ddot{s}_4,\ddot{s}_4])$ | $([\dot{s}_3,\dot{s}_4],[\ddot{s}_1,\ddot{s}_2])$ |

Table 2 The index values with expects four supply chain by expert e2 in the 1st round

| Supply chain | Attribute $(c_1)$ | Attribute $(c_2)$ | Attribute $(c_3)$ | Attribute $(c_4)$ |
|---|---|---|---|---|
| $a_1$ | $([\dot{s}_4,\dot{s}_4],[\ddot{s}_3,\ddot{s}_4])$ | $([\dot{s}_3,\dot{s}_4],[\ddot{s}_2,\ddot{s}_3])$ | $([\dot{s}_3,\dot{s}_4],[\ddot{s}_3,\ddot{s}_3])$ | $([\dot{s}_5,\dot{s}_6],[\ddot{s}_3,\ddot{s}_3])$ |
| $a_2$ | $([\dot{s}_4,\dot{s}_5],[\ddot{s}_3,\ddot{s}_4])$ | $([\dot{s}_2,\dot{s}_3],[\ddot{s}_2,\ddot{s}_3])$ | $([\dot{s}_4,\dot{s}_5],[\ddot{s}_3,\ddot{s}_3])$ | $([\dot{s}_2,\dot{s}_3],[\ddot{s}_3,\ddot{s}_4])$ |
| $a_3$ | $([\dot{s}_3,\dot{s}_4],[\ddot{s}_3,\ddot{s}_4])$ | $([\dot{s}_4,\dot{s}_4],[\ddot{s}_2,\ddot{s}_3])$ | $([\dot{s}_2,\dot{s}_3],[\ddot{s}_3,\ddot{s}_3])$ | $([\dot{s}_3,\dot{s}_4],[\ddot{s}_3,\ddot{s}_4])$ |
| $a_4$ | $([\dot{s}_5,\dot{s}_5],[\ddot{s}_3,\ddot{s}_4])$ | $([\dot{s}_4,\dot{s}_5],[\ddot{s}_2,\ddot{s}_3])$ | $([\dot{s}_1,\dot{s}_2],[\ddot{s}_3,\ddot{s}_3])$ | $([\dot{s}_4,\dot{s}_4],[\ddot{s}_3,\ddot{s}_4])$ |

Table 3 The index values with expects four supply chain by expert e3 in the 1st round

| Supply chain | Attribute $(c_1)$ | Attribute $(c_2)$ | Attribute $(c_3)$ | Attribute $(c_4)$ |
|---|---|---|---|---|
| $a_1$ | $([\dot{s}_5,\dot{s}_5],[\ddot{s}_2,\ddot{s}_3])$ | $([\dot{s}_3,\dot{s}_3],[\ddot{s}_2,\ddot{s}_2])$ | $([\dot{s}_4,\dot{s}_4],[\ddot{s}_3,\ddot{s}_4])$ | $([\dot{s}_4,\dot{s}_5],[\ddot{s}_1,\ddot{s}_1])$ |
| $a_2$ | $([\dot{s}_4,\dot{s}_4],[\ddot{s}_2,\ddot{s}_3])$ | $([\dot{s}_4,\dot{s}_5],[\ddot{s}_2,\ddot{s}_2])$ | $([\dot{s}_1,\dot{s}_2],[\ddot{s}_3,\ddot{s}_4])$ | $([\dot{s}_3,\dot{s}_3],[\ddot{s}_1,\ddot{s}_1])$ |
| $a_3$ | $([\dot{s}_3,\dot{s}_4],[\ddot{s}_2,\ddot{s}_3])$ | $([\dot{s}_5,\dot{s}_5],[\ddot{s}_2,\ddot{s}_2])$ | $([\dot{s}_1,\dot{s}_1],[\ddot{s}_3,\ddot{s}_4])$ | $([\dot{s}_4,\dot{s}_4],[\ddot{s}_1,\ddot{s}_1])$ |
| $a_4$ | $([\dot{s}_2,\dot{s}_3],[\ddot{s}_2,\ddot{s}_3])$ | $([\dot{s}_2,\dot{s}_3],[\ddot{s}_2,\ddot{s}_2])$ | $([\dot{s}_4,\dot{s}_3],[\ddot{s}_3,\ddot{s}_4])$ | $([\dot{s}_4,\dot{s}_5],[\ddot{s}_1,\ddot{s}_1])$ |

**Step 1**: Based on the 2DULGA operator, we calculate the comprehensive value of each alternative, obtain the group decision preference matrix in the 1st round.

**Table 4** The index values with group decision makes in the first round

| Supply chain | Attribute $(c_1)$ | Attribute $(c_2)$ | Attribute $(c_3)$ | Attribute $(c_4)$ |
|---|---|---|---|---|
| $a_1$ | $([\dot{s}_{4.68},\dot{s}_{4.68}],[\ddot{s}_2,\ddot{s}_3])$ | $([\dot{s}_{2.60},\dot{s}_{3.32}],[\ddot{s}_2,\ddot{s}_2])$ | $([\dot{s}_{3.68},\dot{s}_{4.40}],[\ddot{s}_3,\ddot{s}_3])$ | $([\dot{s}_{3.92},\dot{s}_{4.92}],[\ddot{s}_1,\ddot{s}_1])$ |
| $a_2$ | $([\dot{s}_{3.60}, \dot{s}_{4.32}], [\ddot{s}_2, \ddot{s}_3])$ | $([\dot{s}_{3.76}, \dot{s}_{4.36}], [\ddot{s}_2, \ddot{s}_2])$ | $([\dot{s}_{2.76}, \dot{s}_{3.36}], [\ddot{s}_3, \ddot{s}_3])$ | $([\dot{s}_{3.08}, \dot{s}_{3.40}], [\ddot{s}_1, \ddot{s}_1])$ |
| $a_3$ | $([\dot{s}_{2.6}, \dot{s}_{3.60}], [\ddot{s}_2, \ddot{s}_3])$ | $([\dot{s}_{3.88}, \dot{s}_{4.28}], [\ddot{s}_2, \ddot{s}_2])$ | $([\dot{s}_{2.12}, \dot{s}_{2.84}], [\ddot{s}_3, \ddot{s}_3])$ | $([\dot{s}_{3.68}, \dot{s}_{4.40}], [\ddot{s}_1, \ddot{s}_1])$ |
| $a_4$ | $([\dot{s}_{4.16}, \dot{s}_{4.84}], [\ddot{s}_2, \ddot{s}_3])$ | $([\dot{s}_{2.24}, \dot{s}_{3.24}], [\ddot{s}_2, \ddot{s}_2])$ | $([\dot{s}_{2.24}, \dot{s}_{3.24}], [\ddot{s}_3, \ddot{s}_3])$ | $([\dot{s}_{3.60}, \dot{s}_{4.28}], [\ddot{s}_1, \ddot{s}_1])$ |

**Step2:calculate the distance matrix between the individual experts matrix and group decision in the first round**

**Table 5** ,the distance matrix between the individual experts e1 and group decision matrix in the 1-st round

| Supply chain | Attribute $(c_1)$ | Attribute $(c_2)$ | Attribute $(c_3)$ | Attribute $(c_4)$ |
|---|---|---|---|---|
| $a_1$ | $\hat{d}^{11}_{11}=0.033$ | $\hat{d}^{11}_{12}=0.066$ | $\hat{d}^{11}_{13}=0.245$ | $\hat{d}^{11}_{14}=0.073$ |
| $a_2$ | $\hat{d}^{11}_{21}=0.048$ | $\hat{d}^{11}_{22}=0.287$ | $\hat{d}^{11}_{23}=0.118$ | $\hat{d}^{11}_{24}=0.115$ |
| $a_3$ | $\hat{d}^{11}_{31}=0.063$ | $\hat{d}^{11}_{32}=0.098$ | $\hat{d}^{11}_{33}=0.273$ | $\hat{d}^{11}_{34}=0.113$ |
| $a_4$ | $\hat{d}^{11}_{41}=0.104$ | $\hat{d}^{11}_{42}=0.041$ | $\hat{d}^{11}_{43}=0.074$ | $\hat{d}^{11}_{44}=0.073$ |

Table 6 **the distance matrix between the** individual experts e2 and group decision matrix in the 1-st round

| Supply chain | Attribute $(c_1)$ | Attribute $(c_2)$ | Attribute $(c_3)$ | Attribute $(c_4)$ |
|---|---|---|---|---|
| $a_1$ | $\hat{d}^{21}_{11}=0.096$ | $\hat{d}^{21}_{12}=0.118$ | $\hat{d}^{21}_{13}=0.068$ | $\hat{d}^{21}_{14}=0.618$ |
| $a_2$ | $\hat{d}^{21}_{21}=0.244$ | $\hat{d}^{21}_{22}=0.084$ | $\hat{d}^{21}_{23}=0.180$ | $\hat{d}^{21}_{24}=0.230$ |
| $a_3$ | $\hat{d}^{21}_{31}=0.188$ | $\hat{d}^{21}_{32}=0.088$ | $\hat{d}^{21}_{33}=0.018$ | $\hat{d}^{21}_{34}=0.342$ |
| $a_4$ | $\hat{d}^{21}_{41}=0.260$ | $\hat{d}^{21}_{42}=0.240$ | $\hat{d}^{21}_{43}=0.155$ | $\hat{d}^{21}_{44}=0.419$ |

Table 7 **the distance matrix between the** individual experts e3 and group decision matrix in the 1-st round

| Supply chain | Attribute $(c_1)$ | Attribute $(c_2)$ | Attribute $(c_3)$ | Attribute $(c_4)$ |
|---|---|---|---|---|
| $a_1$ | $\hat{d}^{31}_{11}=0.033$ | $\hat{d}^{31}_{12}=0.030$ | $\hat{d}^{31}_{13}=0.103$ | $\hat{d}^{31}_{14}=0.003$ |
| $a_2$ | $\hat{d}^{31}_{21}=0.038$ | $\hat{d}^{31}_{22}=0.037$ | $\hat{d}^{31}_{23}=0.164$ | $\hat{d}^{31}_{24}=0.010$ |
| $a_3$ | $\hat{d}^{31}_{31}=0.042$ | $\hat{d}^{31}_{32}=0.077$ | $\hat{d}^{31}_{33}=0.164$ | $\hat{d}^{31}_{34}=0.015$ |
| $a_4$ | $\hat{d}^{31}_{41}=0.208$ | $\hat{d}^{31}_{42}=0.020$ | $\hat{d}^{31}_{43}=0.314$ | $\hat{d}^{31}_{44}=0.023$ |

Step3:calculate the expectation of two-dimensional language evaluation information $E(\hat{s}^{kt}_{ij})$ according to the follow equation

$$E(\hat{s}_{ij}^{kt}) = \frac{a_{ij}^{kt} + b_{ij}^{kt}}{2*(l-1)} * \frac{c_{ij}^{kt} + d_{ij}^{kt}}{2*(z-1)} \quad (i,j=1,2,3,4; k=1,2,3)$$

Table 8  calculate the expectation value of expert e1 in the 1-st round

| Supply chain | Attribute $(c_1)$ | Attribute $(c_2)$ | Attribute $(c_3)$ | Attribute $(c_4)$ |
|---|---|---|---|---|
| $a_1$ | $E(\hat{s}_{11}^{11})$=0.521 | $E(\hat{s}_{12}^{11})$=0.313 | $E(\hat{s}_{13}^{11})$=0.750 | $E(\hat{s}_{14}^{11})$=0.219 |
| $a_2$ | $E(\hat{s}_{21}^{11})$=0.365 | $E(\hat{s}_{22}^{11})$=0.625 | $E(\hat{s}_{23}^{11})$=0.500 | $E(\hat{s}_{24}^{11})$=0.250 |
| $a_3$ | $E(\hat{s}_{31}^{11})$=0.260 | $E(\hat{s}_{32}^{11})$=0.438 | $E(\hat{s}_{33}^{11})$=0.583 | $E(\hat{s}_{34}^{11})$=0.281 |
| $a_4$ | $E(\hat{s}_{41}^{11})$=0.573 | $E(\hat{s}_{42}^{11})$=0.188 | $E(\hat{s}_{43}^{11})$=0.417 | $E(\hat{s}_{44}^{11})$=0.219 |

Table 9  calculate the expectation value of expert e2 in the 1-st round

| Supply chain | Attribute $(c_1)$ | Attribute $(c_2)$ | Attribute $(c_3)$ | Attribute $(c_4)$ |
|---|---|---|---|---|
| $a_1$ | $E(\hat{s}_{11}^{21})$=0.583 | $E(\hat{s}_{12}^{21})$=0.365 | $E(\hat{s}_{13}^{21})$=0.438 | $E(\hat{s}_{14}^{21})$=0.802 |
| $a_2$ | $E(\hat{s}_{21}^{21})$=0.656 | $E(\hat{s}_{22}^{21})$=0.260 | $E(\hat{s}_{23}^{21})$=0.563 | $E(\hat{s}_{24}^{21})$=0.365 |
| $a_3$ | $E(\hat{s}_{31}^{21})$=0.510 | $E(\hat{s}_{32}^{21})$=0.417 | $E(\hat{s}_{33}^{21})$=0.313 | $E(\hat{s}_{34}^{21})$=0.510 |
| $a_4$ | $E(\hat{s}_{41}^{21})$=0.729 | $E(\hat{s}_{42}^{21})$=0.469 | $E(\hat{s}_{43}^{21})$=0.188 | $E(\hat{s}_{44}^{21})$=0.583 |

Table 10 the expectation value of expert e3 in the 1-st round

| Supply chain | Attribute $(c_1)$ | Attribute $(c_2)$ | Attribute $(c_3)$ | Attribute $(c_4)$ |
|---|---|---|---|---|
| $a_1$ | $E(\hat{s}_{11}^{31})$=0.521 | $E(\hat{s}_{12}^{31})$=0.250 | $E(\hat{s}_{13}^{31})$=0.583 | $E(\hat{s}_{14}^{31})$=0.188 |
| $a_2$ | $E(\hat{s}_{21}^{31})$=0.417 | $E(\hat{s}_{22}^{31})$=0.375 | $E(\hat{s}_{23}^{31})$=0.219 | $E(\hat{s}_{24}^{31})$=0.125 |
| $a_3$ | $E(\hat{s}_{31}^{31})$=0.365 | $E(\hat{s}_{32}^{31})$=0.417 | $E(\hat{s}_{33}^{31})$=0.146 | $E(\hat{s}_{34}^{31})$=0.167 |
| $a_4$ | $E(\hat{s}_{41}^{31})$=0.260 | $E(\hat{s}_{42}^{31})$=0.208 | $E(\hat{s}_{43}^{31})$=0.656 | $E(\hat{s}_{44}^{31})$=0.188 |

**Step 4**: calculate the group consistency index according to the following equation to obtain the vector of group consistency index according to the following equation

$$\eta_k^t = \sum_{i=1}^{m}\sum_{j=1}^{n} d(\hat{s}_{ij}^{kt}, \hat{y}_{ij}^{kt}) / E(\hat{s}_{ij}^{kt})$$

$\eta_1 = (4.464, 6.776, 4.408)$

**Step5**: Determine whether to perform group interaction, Determine whether the group consistency vector index satisfies the condition, and $\eta_{11} \succ 1.5, \eta_{12} \succ 2.5, \eta_{13} \succ 1.5$, hence we need to perform group interaction。

**Step 6:** Calculate the entropy weight under each attribute according to the equation

Table 11   entropy weight under each attribute in the first round

| Supply chain | Attribute $(c_1)$ | Attribute $(c_2)$ | Attribute $(c_3)$ | Attribute $(c_4)$ |
|---|---|---|---|---|
| $a_1$ | $I_{11}^1 = 0.999$ | $I_{12}^1 = 0.989$ | $I_{13}^1 = 0.979$ | $I_{14}^1 = 0.972$ |
| $a_2$ | $I_{21}^1 = 0.963$ | $I_{22}^1 = 0.941$ | $I_{23}^1 = 0.938$ | $I_{24}^1 = 0.925$ |
| $a_3$ | $I_{31}^1 = 0.967$ | $I_{32}^1 = 1.000$ | $I_{33}^1 = 0.875$ | $I_{34}^1 = 0.910$ |
| $a_4$ | $I_{41}^1 = 0.930$ | $I_{42}^1 = 0.916$ | $I_{43}^1 = 0.900$ | $I_{44}^1 = 0.874$ |
| Average information | $I_1^1 = 0.966$ | $I_2^1 = 0.962$ | $I_3^1 = 0.923$ | $I_4^1 = 0.875$ |
| entropy weight | $w_1^1 = 0.259$ | $w_2^1 = 0.258$ | $w_3^1 = 0.248$ | $w_4^1 = 0.235$ |

Step 7: Adjust decision makers' preference information according to the equation, then obtain the new round of decision makers' individual decision preference information, return to step 1.

**The Second round of group decision**

Table 12   The index values with expects four supply chain by experts e1 in the 2$^{nd}$ round

| Supply chain | Attribute $(c_1)$ | Attribute $(c_2)$ | Attribute $(c_3)$ | Attribute $(c_4)$ |
|---|---|---|---|---|
| $a_1$ | $([\dot{s}_{4.917}, \dot{s}_{4.917}], [\ddot{s}_2, \ddot{s}_3])$ | $([\dot{s}_{2.155}, \dot{s}_{3.083}], [\ddot{s}_2, \ddot{s}_2])$ | $([\dot{s}_{3.921}, \dot{s}_{4.851}], [\ddot{s}_3, \ddot{s}_3])$ | $([\dot{s}_{3.216}, \dot{s}_{4.216}], [\ddot{s}_1, \ddot{s}_1])$ |
| $a_2$ | $([\dot{s}_{3.156}, \dot{s}_{4.083}], [\ddot{s}_2, \ddot{s}_3])$ | $([\dot{s}_{4.680}, \dot{s}_{4.835}], [\ddot{s}_2, \ddot{s}_2])$ | $([\dot{s}_{2.941}, \dot{s}_{3.089}], [\ddot{s}_3, \ddot{s}_3])$ | $([\dot{s}_{3.784}, \dot{s}_{3.859}], [\ddot{s}_1, \ddot{s}_1])$ |
| $a_3$ | $([\dot{s}_{2.156}, \dot{s}_{3.156}], [\ddot{s}_2, \ddot{s}_3])$ | $([\dot{s}_{3.227}, \dot{s}_{4.072}], [\ddot{s}_2, \ddot{s}_2])$ | $([\dot{s}_{2.782}, \dot{s}_{3.713}], [\ddot{s}_3, \ddot{s}_3])$ | $([\dot{s}_{3.925}, \dot{s}_{4.859}], [\ddot{s}_1, \ddot{s}_1])$ |
| $a_4$ | $([\dot{s}_{4.782}, \dot{s}_{5.699}], [\ddot{s}_2, \ddot{s}_3])$ | $([\dot{s}_{1.320}, \dot{s}_{2.320}], [\ddot{s}_2, \ddot{s}_2])$ | $([\dot{s}_{2.059}, \dot{s}_{3.059}], [\ddot{s}_3, \ddot{s}_3])$ | $([\dot{s}_{3.141}, \dot{s}_{4.056}], [\ddot{s}_1, \ddot{s}_1])$ |

Table 13   The index values with expects four supply chain by experts e2 in the 2$^{nd}$ round

| Supply chain | Attribute $(c_1)$ | Attribute $(c_2)$ | Attribute $(c_3)$ | Attribute $(c_4)$ |
|---|---|---|---|---|
| $a_1$ | $([\dot{s}_{4.176}, \dot{s}_{4.176}], [\ddot{s}_2, \ddot{s}_3])$ | $([\dot{s}_{2.897}, \dot{s}_{3.825}], [\ddot{s}_2, \ddot{s}_2])$ | $([\dot{s}_{3.168}, \dot{s}_{4.099}], [\ddot{s}_3, \ddot{s}_3])$ | $([\dot{s}_{4.746}, \dot{s}_{5.746}], [\ddot{s}_1, \ddot{s}_1])$ |
| $a_2$ | $([\dot{s}_{3.896}, \dot{s}_{4.824}], [\ddot{s}_2, \ddot{s}_3])$ | $([\dot{s}_{2.454}, \dot{s}_{3.351}], [\ddot{s}_2, \ddot{s}_2])$ | $([\dot{s}_{3.693}, \dot{s}_{4.594}], [\ddot{s}_3, \ddot{s}_3])$ | $([\dot{s}_{2.254}, \dot{s}_{3.094}], [\ddot{s}_1, \ddot{s}_1])$ |
| $a_3$ | $([\dot{s}_{2.896}, \dot{s}_{3.896}], [\ddot{s}_2, \ddot{s}_3])$ | $([\dot{s}_{3.969}, \dot{s}_{4.072}], [\ddot{s}_2, \ddot{s}_2])$ | $([\dot{s}_{2.030}, \dot{s}_{2.690}], [\ddot{s}_3, \ddot{s}_3])$ | $([\dot{s}_{3.160}, \dot{s}_{4.094}], [\ddot{s}_1, \ddot{s}_1])$ |
| $a_4$ | $([\dot{s}_{4.782}, \dot{s}_{4.959}], [\ddot{s}_2, \ddot{s}_3])$ | $([\dot{s}_{3.546}, \dot{s}_{4.456}], [\ddot{s}_2, \ddot{s}_2])$ | $([\dot{s}_{1.307}, \dot{s}_{2.307}], [\ddot{s}_3, \ddot{s}_3])$ | $([\dot{s}_{3.906}, \dot{s}_{4.066}], [\ddot{s}_1, \ddot{s}_1])$ |

Table 14   The index values with expects four supply chain by experts e3 in the 2nd round

| Supply chain | Attribute $(c_1)$ | Attribute $(c_2)$ | Attribute $(c_3)$ | Attribute $(c_4)$ |
|---|---|---|---|---|
| $a_1$ | $([\dot{s}_{4.917}, \dot{s}_{4.917}], [\ddot{s}_2, \ddot{s}_3])$ | $([\dot{s}_{2.897}, \dot{s}_{3.083}], [\ddot{s}_2, \ddot{s}_2])$ | $([\dot{s}_{3.921}, \dot{s}_{4.099}], [\ddot{s}_3, \ddot{s}_3])$ | $([\dot{s}_{3.981}, \dot{s}_{4.981}], [\ddot{s}_1, \ddot{s}_1])$ |
| $a_2$ | $([\dot{s}_{3.896}, \dot{s}_{4.083}],[\ddot{s}_2,\ddot{s}_3])$ | $([\dot{s}_{3.938}, \dot{s}_{4.835}],[\ddot{s}_2,\ddot{s}_2])$ | $([\dot{s}_{1.436}, \dot{s}_{2.337}],[\ddot{s}_3,\ddot{s}_3])$ | $([\dot{s}_{3.019}, \dot{s}_{3.094}],[\ddot{s}_1,\ddot{s}_1])$ |
| $a_3$ | $([\dot{s}_{2.896}, \dot{s}_{3.896}],[\ddot{s}_2,\ddot{s}_3])$ | $([\dot{s}_{4.711}, \dot{s}_{4.184}],[\ddot{s}_2,\ddot{s}_2])$ | $([\dot{s}_{1.277}, \dot{s}_{1.456}],[\ddot{s}_3,\ddot{s}_3])$ | $([\dot{s}_{3.925}, \dot{s}_{4.094}],[\ddot{s}_1,\ddot{s}_1])$ |
| $a_4$ | $([\dot{s}_{2.560}, \dot{s}_{3.477}],[\ddot{s}_2,\ddot{s}_3])$ | $([\dot{s}_{2.062}, \dot{s}_{3.062}],[\ddot{s}_2,\ddot{s}_2])$ | $([\dot{s}_{3.564}, \dot{s}_{4.564}],[\ddot{s}_3,\ddot{s}_3])$ | $([\dot{s}_{3.096}, \dot{s}_{4.831}],[\ddot{s}_1,\ddot{s}_1])$ |

**Step 1**:Based on the 2DULGA operator,we calculate the comprehensive value of each alternative,obtain the group decision preference matrix in the 2nd round

Table15 The index values with group decision makes in the 2nd round

| Supply chain | Attribute $(c_1)$ | Attribute $(c_2)$ | Attribute $(c_3)$ | Attribute $(c_4)$ |
|---|---|---|---|---|
| $a_1$ | $([\dot{s}_{4.670}, \dot{s}_{4.670}],[\ddot{s}_2,\ddot{s}_3])$ | $([\dot{s}_{2.649}, \dot{s}_{3.330}],[\ddot{s}_2,\ddot{s}_2])$ | $([\dot{s}_{3.670}, \dot{s}_{4.350}],[\ddot{s}_3,\ddot{s}_3])$ | $([\dot{s}_{3.981}, \dot{s}_{4.981}],[\ddot{s}_1,\ddot{s}_1])$ |
| $a_2$ | $([\dot{s}_{3.649}, \dot{s}_{4.330}],[\ddot{s}_2,\ddot{s}_3])$ | $([\dot{s}_{3.691}, \dot{s}_{4.340}],[\ddot{s}_2,\ddot{s}_2])$ | $([\dot{s}_{2.690}, \dot{s}_{3.340}],[\ddot{s}_3,\ddot{s}_3])$ | $([\dot{s}_{3.019}, \dot{s}_{3.349}],[\ddot{s}_1,\ddot{s}_1])$ |
| $a_3$ | $([\dot{s}_{2.649}, \dot{s}_{3.649}],[\ddot{s}_2,\ddot{s}_3])$ | $([\dot{s}_{3.969}, \dot{s}_{4.320}],[\ddot{s}_2,\ddot{s}_2])$ | $([\dot{s}_{2.030}, \dot{s}_{2.710}],[\ddot{s}_3,\ddot{s}_3])$ | $([\dot{s}_{3.670}, \dot{s}_{4.349}],[\ddot{s}_1,\ddot{s}_1])$ |
| $a_4$ | $([\dot{s}_{4.041}, \dot{s}_{4.712}],[\ddot{s}_2,\ddot{s}_3])$ | $([\dot{s}_{2.309}, \dot{s}_{3.309}],[\ddot{s}_2,\ddot{s}_2])$ | $([\dot{s}_{2.310}, \dot{s}_{3.310}],[\ddot{s}_3,\ddot{s}_3])$ | $([\dot{s}_{3.651}, \dot{s}_{4.321}],[\ddot{s}_1,\ddot{s}_1])$ |

**Step2**:calculate the distance matrix between the individual experts matrix and group decision in the second round

Table 16 **the distance matrix between the** individual experts e1 and group decision in the 2-nd round

| Supply chain | Attribute $(c_1)$ | Attribute $(c_2)$ | Attribute $(c_3)$ | Attribute $(c_4)$ |
|---|---|---|---|---|
| $a_1$ | $\hat{d}_{11}^{12}=0.026$ | $\hat{d}_{12}^{12}=0.031$ | $\hat{d}_{13}^{12}=0.047$ | $\hat{d}_{14}^{12}=0.032$ |
| $a_2$ | $\hat{d}_{21}^{12}=0.039$ | $\hat{d}_{22}^{12}=0.062$ | $\hat{d}_{23}^{12}=0.031$ | $\hat{d}_{24}^{12}=0.027$ |
| $a_3$ | $\hat{d}_{31}^{12}=0.051$ | $\hat{d}_{32}^{12}=0.041$ | $\hat{d}_{33}^{12}=0.110$ | $\hat{d}_{34}^{12}=0.016$ |
| $a_4$ | $\hat{d}_{41}^{12}=0.090$ | $\hat{d}_{42}^{12}=0.082$ | $\hat{d}_{43}^{12}=0.031$ | $\hat{d}_{44}^{12}=0.016$ |

Table 17 **the distance matrix between the** individual experts e2 and group decision in the 2-nd round

| Supply chain | Attribute $(c_1)$ | Attribute $(c_2)$ | Attribute $(c_3)$ | Attribute $(c_4)$ |
|---|---|---|---|---|
| $a_1$ | $\hat{d}_{11}^{22}=0.051$ | $\hat{d}_{12}^{22}=0.031$ | $\hat{d}_{13}^{22}=0.047$ | $\hat{d}_{14}^{22}=0.032$ |
| $a_2$ | $\hat{d}_{21}^{22}=0.039$ | $\hat{d}_{22}^{22}=0.093$ | $\hat{d}_{23}^{22}=0.141$ | $\hat{d}_{24}^{22}=0.021$ |
| $a_3$ | $\hat{d}_{31}^{22}=0.026$ | $\hat{d}_{32}^{22}=0.010$ | $\hat{d}_{33}^{22}=0.016$ | $\hat{d}_{34}^{22}=0.016$ |
| $a_4$ | $\hat{d}_{41}^{22}=0.051$ | $\hat{d}_{42}^{22}=0.103$ | $\hat{d}_{43}^{22}=0.125$ | $\hat{d}_{44}^{22}=0.011$ |

Table 18 **the distance matrix between the** individual experts e3 and group decision in the 2-nd

round

| Supply chain | Attribute $(c_1)$ | Attribute $(c_2)$ | Attribute $(c_3)$ | Attribute $(c_4)$ |
|---|---|---|---|---|
| $a_1$ | $\hat{d}_{11}^{32}=0.026$ | $\hat{d}_{12}^{32}=0.021$ | $\hat{d}_{13}^{32}=0.031$ | $\hat{d}_{14}^{32}=0.000$ |
| $a_2$ | $\hat{d}_{21}^{32}=0.026$ | $\hat{d}_{22}^{32}=0.031$ | $\hat{d}_{23}^{32}=0.141$ | $\hat{d}_{24}^{32}=0.005$ |
| $a_3$ | $\hat{d}_{31}^{32}=0.026$ | $\hat{d}_{32}^{32}=0.052$ | $\hat{d}_{33}^{32}=0.125$ | $\hat{d}_{34}^{32}=0.011$ |
| $a_4$ | $\hat{d}_{41}^{32}=0.141$ | $\hat{d}_{42}^{32}=0.021$ | $\hat{d}_{43}^{32}=0.157$ | $\hat{d}_{44}^{32}=0.016$ |

Step3: calculate the expectation of two-dimensional language evaluation information $E(\hat{s}_{ij}^{kt})$ according to the follow equation

$$E(\hat{s}_{ij}^{kt})=\frac{a_{ij}^{kt}+b_{ij}^{kt}}{2*(l-1)}*\frac{c_{ij}^{kt}+d_{ij}^{kt}}{2*(z-1)} \ (i,j=1,2,3,4;k=1,2,3)$$

Table 19  the expectation value of expert e1 in the 2-nd round

| Supply chain | Attribute $(c_1)$ | Attribute $(c_2)$ | Attribute $(c_3)$ | Attribute $(c_4)$ |
|---|---|---|---|---|
| $a_1$ | $E(\hat{s}_{11}^{12})=0.512$ | $E(\hat{s}_{12}^{12})=0.218$ | $E(\hat{s}_{13}^{12})=0.548$ | $E(\hat{s}_{14}^{12})=0.155$ |
| $a_2$ | $E(\hat{s}_{21}^{12})=0.377$ | $E(\hat{s}_{22}^{12})=0.396$ | $E(\hat{s}_{23}^{12})=0.377$ | $E(\hat{s}_{24}^{12})=0.159$ |
| $a_3$ | $E(\hat{s}_{31}^{12})=0.277$ | $E(\hat{s}_{32}^{12})=0.304$ | $E(\hat{s}_{33}^{12})=0.406$ | $E(\hat{s}_{34}^{12})=0.183$ |
| $a_4$ | $E(\hat{s}_{41}^{12})=0.546$ | $E(\hat{s}_{42}^{12})=0.152$ | $E(\hat{s}_{43}^{12})=0.320$ | $E(\hat{s}_{44}^{12})=0.150$ |

Table 20  the expectation value of expert e2 in the 2-nd round

| Supply chain | Attribute $(c_1)$ | Attribute $(c_2)$ | Attribute $(c_3)$ | Attribute $(c_4)$ |
|---|---|---|---|---|
| $a_1$ | $E(\hat{s}_{11}^{22})=0.435$ | $E(\hat{s}_{12}^{22})=0.280$ | $E(\hat{s}_{13}^{22})=0.454$ | $E(\hat{s}_{14}^{22})=0.219$ |
| $a_2$ | $E(\hat{s}_{21}^{22})=0.454$ | $E(\hat{s}_{22}^{22})=0.242$ | $E(\hat{s}_{23}^{22})=0.518$ | $E(\hat{s}_{24}^{22})=0.111$ |
| $a_3$ | $E(\hat{s}_{31}^{22})=0.354$ | $E(\hat{s}_{32}^{22})=0.335$ | $E(\hat{s}_{33}^{22})=0.312$ | $E(\hat{s}_{34}^{22})=0.151$ |
| $a_4$ | $E(\hat{s}_{41}^{22})=0.507$ | $E(\hat{s}_{42}^{22})=0.337$ | $E(\hat{s}_{43}^{22})=0.226$ | $E(\hat{s}_{44}^{22})=0.150$ |

Table 21  the expectation value of expert e2 in the 2-nd round

| Supply chain | Attribute $(c_1)$ | Attribute $(c_2)$ | Attribute $(c_3)$ | Attribute $(c_4)$ |
|---|---|---|---|---|
| $a_1$ | $E(\hat{s}_{11}^{31})=0.512$ | $E(\hat{s}_{12}^{31})=0.249$ | $E(\hat{s}_{13}^{31})=0.501$ | $E(\hat{s}_{14}^{31})=0.187$ |
| $a_2$ | $E(\hat{s}_{21}^{31})=0.416$ | $E(\hat{s}_{22}^{31})=0.366$ | $E(\hat{s}_{23}^{31})=0.236$ | $E(\hat{s}_{24}^{31})=0.127$ |
| $a_3$ | $E(\hat{s}_{31}^{31})=0.354$ | $E(\hat{s}_{32}^{31})=0.397$ | $E(\hat{s}_{33}^{31})=0.171$ | $E(\hat{s}_{34}^{31})=0.167$ |
| $a_4$ | $E(\hat{s}_{41}^{31})=0.314$ | $E(\hat{s}_{42}^{31})=0.213$ | $E(\hat{s}_{43}^{31})=0.508$ | $E(\hat{s}_{44}^{31})=0.182$ |

**Step 4:** calculate the group consistency index according to the following equation to obtain the vector of group consistency index

$$\eta_k^t = \sum_{i=1}^{m}\sum_{j=1}^{n} d(\hat{s}_{ij}^{kt}, \hat{y}_{ij}^{kt})/ E(\hat{s}_{ij}^{kt})$$

$$\eta_2 = (2.583, 3.382, 2.924)$$

**Step5:** Determine whether to perform group interaction, Determine whether the group consistency vector index satisfies the condition, and $\eta_{11} \succ 1.5, \eta_{12} \succ 2.5, \eta_{13} \succ 1.5$, hence we need to perform group interaction.

**Step6:** Calculate the entropy weight under each attribute according to the equation

Table 22 entropy weight under each attribute in the second round

| Supply chain | Attribute $(c_1)$ | Attribute $(c_2)$ | Attribute $(c_3)$ | Attribute $(c_4)$ |
|---|---|---|---|---|
| $a_1$ | $I_{11}^2=0.997$ | $I_{12}^2=0.995$ | $I_{13}^2=0.997$ | $I_{14}^2=0.991$ |
| $a_2$ | $I_{21}^2=0.997$ | $I_{22}^2=0.981$ | $I_{23}^2=0.956$ | $I_{24}^2=0.990$ |
| $a_3$ | $I_{31}^2=0.994$ | $I_{32}^2=0.994$ | $I_{33}^2=0.949$ | $I_{34}^2=0.997$ |
| $a_4$ | $I_{41}^2=0.976$ | $I_{42}^2=0.951$ | $I_{43}^2=0.950$ | $I_{44}^2=0.997$ |
| Average information entropy | $I_1^2=0.991$ | $I_2^2=0.981$ | $I_3^2=0.963$ | $I_4^2=0.994$ |
| entropy weight | $w_1^2=0.252$ | $w_2^2=0.250$ | $w_3^2=0.245$ | $w_4^2=0.253$ |

Step 7: Adjust decision makers' preference information according to the equation, then obtain ta new round of decision makers' individual decision preference information.

**The Third round of group decision**

Table 23 The index values with expects four supply chain by experts e1 in the 3rdround

| Supply chain | Attribute $(c_1)$ | Attribute $(c_2)$ | Attribute $(c_3)$ | Attribute $(c_4)$ |
|---|---|---|---|---|
| $a_1$ | $([\dot{s}_{4.855}, \dot{s}_{4.855}], [\ddot{s}_2, \ddot{s}_3])$ | $([\dot{s}_{2.278}, \dot{s}_{3.144}], [\ddot{s}_2, \ddot{s}_2])$ | $([\dot{s}_{3.859}, \dot{s}_{4.728}], [\ddot{s}_3, \ddot{s}_3])$ | $([\dot{s}_{3.469}, \dot{s}_{4.410}], [\ddot{s}_1, \ddot{s}_1])$ |
| $a_2$ | $([\dot{s}_{3.280}, \dot{s}_{4.145}], [\ddot{s}_2, \ddot{s}_3])$ | $([\dot{s}_{3.433}, \dot{s}_{4.711}], [\ddot{s}_2, \ddot{s}_2])$ | $([\dot{s}_{2.879}, \dot{s}_{3.151}], [\ddot{s}_3, \ddot{s}_3])$ | $([\dot{s}_{3.803}, \dot{s}_{3.730}], [\ddot{s}_1, \ddot{s}_1])$ |
| $a_3$ | $([\dot{s}_{2.280}, \dot{s}_{3.280}], [\ddot{s}_2, \ddot{s}_3])$ | $([\dot{s}_{3.412}, \dot{s}_{4.134}], [\ddot{s}_2, \ddot{s}_2])$ | $([\dot{s}_{2.598}, \dot{s}_{3.467}], [\ddot{s}_3, \ddot{s}_3])$ | $([\dot{s}_{4.161}, \dot{s}_{4.730}], [\ddot{s}_1, \ddot{s}_1])$ |
| $a_4$ | $([\dot{s}_{4.595}, \dot{s}_{5.450}], [\ddot{s}_2, \ddot{s}_3])$ | $([\dot{s}_{1.567}, \dot{s}_{2.567}], [\ddot{s}_2, \ddot{s}_2])$ | $([\dot{s}_{2.121}, \dot{s}_{3.121}], [\ddot{s}_3, \ddot{s}_3])$ | $([\dot{s}_{3.375}, \dot{s}_{4.130}], [\ddot{s}_1, \ddot{s}_1])$ |

Table 24 The index values with expects four supply chain by experts e2 in the 3rd round

| Supplychain | Attribute $(c_1)$ | Attribute $(c_2)$ | Attribute $(c_3)$ | Attribute $(c_4)$ |
|---|---|---|---|---|
| $a_1$ | $([\dot{s}_{4.301}, \dot{s}_{4.301}], [\ddot{s}_2, \ddot{s}_3])$ | $([\dot{s}_{2.835}, \dot{s}_{3.701}], [\ddot{s}_2, \ddot{s}_2])$ | $([\dot{s}_{3.291}, \dot{s}_{4.161}], [\ddot{s}_3, \ddot{s}_3])$ | $([\dot{s}_{4.553}, \dot{s}_{5.553}], [\ddot{s}_1, \ddot{s}_1])$ |
| $a_2$ | $([\dot{s}_{3.384}, \dot{s}_{4.699}],[\ddot{s}_2, \ddot{s}_3])$ | $([\dot{s}_{2.763}, \dot{s}_{3.598}],[\ddot{s}_2, \ddot{s}_2])$ | $([\dot{s}_{3.447}, \dot{s}_{4.286}],[\ddot{s}_3, \ddot{s}_3])$ | $([\dot{s}_{2.447}, \dot{s}_{3.158}],[\ddot{s}_1, \ddot{s}_1])$ |
| $a_3$ | $([\dot{s}_{2.834}, \dot{s}_{3.834}],[\ddot{s}_2, \ddot{s}_3])$ | $([\dot{s}_{3.969}, \dot{s}_{4.134}],[\ddot{s}_2, \ddot{s}_2])$ | $([\dot{s}_{2.030}, \dot{s}_{2.899}],[\ddot{s}_3, \ddot{s}_3])$ | $([\dot{s}_{3.289}, \dot{s}_{4.158}],[\ddot{s}_1, \ddot{s}_1])$ |
| $a_4$ | $([\dot{s}_{4.595}, \dot{s}_{4.896}],[\ddot{s}_2, \ddot{s}_3])$ | $([\dot{s}_{3.237}, \dot{s}_{4.237}],[\ddot{s}_2, \ddot{s}_2])$ | $([\dot{s}_{1.553}, \dot{s}_{2.553}],[\ddot{s}_3, \ddot{s}_3])$ | $([\dot{s}_{3.842}, \dot{s}_{4.130}],[\ddot{s}_1, \ddot{s}_1])$ |

Table 25   The index values with expects four supply chain by experts e3 in the third round

| Supply chain | Attribute $(c_1)$ | Attribute $(c_2)$ | Attribute $(c_3)$ | Attribute $(c_4)$ |
|---|---|---|---|---|
| $a_1$ | $([\dot{s}_{4.855}, \dot{s}_{4.855}],[\ddot{s}_2, \ddot{s}_3])$ | $([\dot{s}_{2.835}, \dot{s}_{3.144}],[\ddot{s}_2, \ddot{s}_2])$ | $([\dot{s}_{3.859}, \dot{s}_{4.161}],[\ddot{s}_3, \ddot{s}_3])$ | $([\dot{s}_{3.981}, \dot{s}_{4.981}],[\ddot{s}_1, \ddot{s}_1])$ |
| $a_2$ | $([\dot{s}_{3.384}, \dot{s}_{4.415}],[\ddot{s}_2, \ddot{s}_3])$ | $([\dot{s}_{3.876}, \dot{s}_{4.711}],[\ddot{s}_2, \ddot{s}_2])$ | $([\dot{s}_{1.743}, \dot{s}_{2.583}],[\ddot{s}_3, \ddot{s}_3])$ | $([\dot{s}_{3.019}, \dot{s}_{3.158}],[\ddot{s}_1, \ddot{s}_1])$ |
| $a_3$ | $([\dot{s}_{2.834}, \dot{s}_{3.834}],[\ddot{s}_2, \ddot{s}_3])$ | $([\dot{s}_{4.526}, \dot{s}_{4.691}],[\ddot{s}_2, \ddot{s}_2])$ | $([\dot{s}_{1.462}, \dot{s}_{1.763}],[\ddot{s}_3, \ddot{s}_3])$ | $([\dot{s}_{3.860}, \dot{s}_{4.158}],[\ddot{s}_1, \ddot{s}_1])$ |
| $a_4$ | $([\dot{s}_{2.934}, \dot{s}_{3.789}],[\ddot{s}_2, \ddot{s}_3])$ | $([\dot{s}_{2.124}, \dot{s}_{3.124}],[\ddot{s}_2, \ddot{s}_2])$ | $([\dot{s}_{3.257}, \dot{s}_{4.257}],[\ddot{s}_3, \ddot{s}_3])$ | $([\dot{s}_{3.842}, \dot{s}_{4.072}],[\ddot{s}_1, \ddot{s}_1])$ |

**Step 1**: Based on the 2DULGA operator, we calculate the comprehensive value of each alternative, obtain the group decision preference matrix in the third round.

Table25   The index values with group decision makes in the third round

| Supply chain | Attribute $(c_1)$ | Attribute $(c_2)$ | Attribute $(c_3)$ | Attribute $(c_4)$ |
|---|---|---|---|---|
| $a_1$ | $([\dot{s}_{4.670}, \dot{s}_{4.670}],[\ddot{s}_2, \ddot{s}_3])$ | $([\dot{s}_{2.649}, \dot{s}_{3.330}],[\ddot{s}_2, \ddot{s}_2])$ | $([\dot{s}_{3.670}, \dot{s}_{4.350}],[\ddot{s}_3, \ddot{s}_3])$ | $([\dot{s}_{4.001}, \dot{s}_{4.981}],[\ddot{s}_1, \ddot{s}_1])$ |
| $a_2$ | $([\dot{s}_{3.649}, \dot{s}_{4.330}],[\ddot{s}_2, \ddot{s}_3])$ | $([\dot{s}_{3.691}, \dot{s}_{4.340}],[\ddot{s}_2, \ddot{s}_2])$ | $([\dot{s}_{2.690}, \dot{s}_{3.340}],[\ddot{s}_3, \ddot{s}_3])$ | $([\dot{s}_{3.090}, \dot{s}_{3.349}],[\ddot{s}_1, \ddot{s}_1])$ |
| $a_3$ | $([\dot{s}_{2.649}, \dot{s}_{3.649}],[\ddot{s}_2, \ddot{s}_3])$ | $([\dot{s}_{3.969}, \dot{s}_{4.320}],[\ddot{s}_2, \ddot{s}_2])$ | $([\dot{s}_{2.030}, \dot{s}_{2.710}],[\ddot{s}_3, \ddot{s}_3])$ | $([\dot{s}_{3.770}, \dot{s}_{4.349}],[\ddot{s}_1, \ddot{s}_1])$ |
| $a_4$ | $([\dot{s}_{4.041}, \dot{s}_{4.712}],[\ddot{s}_2, \ddot{s}_3])$ | $([\dot{s}_{2.309}, \dot{s}_{3.309}],[\ddot{s}_2, \ddot{s}_2])$ | $([\dot{s}_{2.310}, \dot{s}_{3.310}],[\ddot{s}_3, \ddot{s}_3])$ | $([\dot{s}_{3.686}, \dot{s}_{4.321}],[\ddot{s}_1, \ddot{s}_1])$ |

**Step2**: calculate the distance matrix between the individual experts matrix and group decision in the first round

Table 26 **the distance matrix between the** individual experts e1 and group decision in the 3-ed round

| Supply chain | Attribute $(c_1)$ | Attribute $(c_2)$ | Attribute $(c_3)$ | Attribute $(c_4)$ |
|---|---|---|---|---|
| $a_1$ | $\hat{d}_{11}^{13}=0.019$ | $\hat{d}_{12}^{13}=0.023$ | $\hat{d}_{13}^{13}=0.035$ | $\hat{d}_{14}^{13}=0.023$ |
| $a_2$ | $\hat{d}_{21}^{13}=0.029$ | $\hat{d}_{22}^{13}=0.046$ | $\hat{d}_{23}^{13}=0.024$ | $\hat{d}_{24}^{13}=0.023$ |
| $a_3$ | $\hat{d}_{31}^{13}=0.038$ | $\hat{d}_{32}^{13}=0.031$ | $\hat{d}_{33}^{13}=0.083$ | $\hat{d}_{34}^{13}=0.016$ |
| $a_4$ | $\hat{d}_{41}^{13}=0.067$ | $\hat{d}_{42}^{13}=0.062$ | $\hat{d}_{43}^{13}=0.024$ | $\hat{d}_{44}^{13}=0.010$ |

Table 27 **the distance matrix between the** individual experts e2 and group decision in the 3-ed round

| Supply chain | Attribute $(c_1)$ | Attribute $(c_2)$ | Attribute $(c_3)$ | Attribute $(c_4)$ |
|---|---|---|---|---|

| $a_1$ | $\hat{d}_{11}^{23}=0.064$ | $\hat{d}_{12}^{23}=0.023$ | $\hat{d}_{13}^{23}=0.035$ | $\hat{d}_{14}^{23}=0.023$ |
|---|---|---|---|---|
| $a_2$ | $\hat{d}_{21}^{23}=0.067$ | $\hat{d}_{22}^{23}=0.070$ | $\hat{d}_{23}^{23}=0.106$ | $\hat{d}_{24}^{23}=0.017$ |
| $a_3$ | $\hat{d}_{31}^{23}=0.047$ | $\hat{d}_{32}^{23}=0.008$ | $\hat{d}_{33}^{23}=0.012$ | $\hat{d}_{34}^{23}=0.014$ |
| $a_4$ | $\hat{d}_{41}^{23}=0.081$ | $\hat{d}_{42}^{23}=0.077$ | $\hat{d}_{43}^{23}=0.095$ | $\hat{d}_{44}^{23}=0.007$ |

Table28 **the distance matrix between the** individual experts e3 and group decision in the 3-ed round

| Supply chain | Attribute $(c_1)$ | Attribute $(c_2)$ | Attribute $(c_3)$ | Attribute $(c_4)$ |
|---|---|---|---|---|
| $a_1$ | $\hat{d}_{11}^{33}=0.019$ | $\hat{d}_{12}^{33}=0.022$ | $\hat{d}_{13}^{33}=0.024$ | $\hat{d}_{14}^{33}=0.000$ |
| $a_2$ | $\hat{d}_{21}^{33}=0.019$ | $\hat{d}_{22}^{33}=0.025$ | $\hat{d}_{23}^{33}=0.106$ | $\hat{d}_{24}^{33}=0.005$ |
| $a_3$ | $\hat{d}_{31}^{33}=0.019$ | $\hat{d}_{32}^{33}=0.035$ | $\hat{d}_{33}^{33}=0.095$ | $\hat{d}_{34}^{33}=0.006$ |
| $a_4$ | $\hat{d}_{41}^{33}=0.106$ | $\hat{d}_{42}^{33}=0.036$ | $\hat{d}_{43}^{33}=0.118$ | $\hat{d}_{44}^{33}=0.011$ |

Step3：calculate the expectation of two-dimensional language evaluation information $E(\hat{s}_{ij}^{kt})$ according to the follow equation。

$$E(\hat{s}_{ij}^{kt}) = \frac{a_{ij}^{kt}+b_{ij}^{kt}}{2*(l-1)} * \frac{c_{ij}^{kt}+d_{ij}^{kt}}{2*(z-1)} \quad (i,j=1,2,3,4; k=1,2,3)$$

Table29  calculate the expectation value of expert e1 in the 3-ed round

| Supply chain | Attribute $(c_1)$ | Attribute $(c_2)$ | Attribute $(c_3)$ | Attribute $(c_4)$ |
|---|---|---|---|---|
| $a_1$ | $E(\hat{s}_{11}^{13})=0.506$ | $E(\hat{s}_{12}^{13})=0.226$ | $E(\hat{s}_{13}^{13})=0.537$ | $E(\hat{s}_{14}^{13})=0.164$ |
| $a_2$ | $E(\hat{s}_{21}^{13})=0.387$ | $E(\hat{s}_{22}^{13})=0.381$ | $E(\hat{s}_{23}^{13})=0.337$ | $E(\hat{s}_{24}^{13})=0.159$ |
| $a_3$ | $E(\hat{s}_{31}^{13})=0.290$ | $E(\hat{s}_{32}^{13})=0.314$ | $E(\hat{s}_{33}^{13})=0.379$ | $E(\hat{s}_{34}^{13})=0.185$ |
| $a_4$ | $E(\hat{s}_{41}^{13})=0.523$ | $E(\hat{s}_{42}^{13})=0.172$ | $E(\hat{s}_{43}^{13})=0.328$ | $E(\hat{s}_{44}^{13})=0.156$ |

Table 30  calculate the expectation value of expert e2 in the 3-ed round

| Supply chain | Attribute $(c_1)$ | Attribute $(c_2)$ | Attribute $(c_3)$ | Attribute $(c_4)$ |
|---|---|---|---|---|
| $a_1$ | $E(\hat{s}_{11}^{23})=0.448$ | $E(\hat{s}_{12}^{23})=0.272$ | $E(\hat{s}_{13}^{23})=0.466$ | $E(\hat{s}_{14}^{23})=0.211$ |
| $a_2$ | $E(\hat{s}_{21}^{23})=0.444$ | $E(\hat{s}_{22}^{23})=0.265$ | $E(\hat{s}_{23}^{23})=0.483$ | $E(\hat{s}_{24}^{23})=0.117$ |
| $a_3$ | $E(\hat{s}_{31}^{23})=0.347$ | $E(\hat{s}_{32}^{23})=0.338$ | $E(\hat{s}_{33}^{23})=0.308$ | $E(\hat{s}_{34}^{23})=0.155$ |
| $a_4$ | $E(\hat{s}_{41}^{23})=0.494$ | $E(\hat{s}_{42}^{23})=0.311$ | $E(\hat{s}_{43}^{23})=0.257$ | $E(\hat{s}_{44}^{23})=0.166$ |

Table 31  calculate the expectation value of expert e3 in the 3-ed round

| Supply chain | Attribute ($c_1$) | Attribute ($c_2$) | Attribute ($c_3$) | Attribute ($c_4$) |
|---|---|---|---|---|
| $a_1$ | $E(\hat{s}_{11}^{33})=0.506$ | $E(\hat{s}_{12}^{33})=0.249$ | $E(\hat{s}_{13}^{33})=0.501$ | $E(\hat{s}_{14}^{33})=0.187$ |
| $a_2$ | $E(\hat{s}_{21}^{33})=0.416$ | $E(\hat{s}_{22}^{33})=0.358$ | $E(\hat{s}_{23}^{33})=0.270$ | $E(\hat{s}_{24}^{33})=0.129$ |
| $a_3$ | $E(\hat{s}_{31}^{33})=0.347$ | $E(\hat{s}_{32}^{33})=0.384$ | $E(\hat{s}_{33}^{33})=0.202$ | $E(\hat{s}_{34}^{33})=0.167$ |
| $a_4$ | $E(\hat{s}_{41}^{33})=0.350$ | $E(\hat{s}_{42}^{33})=0.219$ | $E(\hat{s}_{43}^{33})=0.470$ | $E(\hat{s}_{44}^{33})=0.178$ |

**Step 4**: calculate the group consistency index according to the following equation to obtain the vector of group consistency index

$$\eta_k^t = \sum_{i=1}^{m}\sum_{j=1}^{n} d(\hat{s}_{ij}^{kt}, \hat{y}_{ij}^{kt}) / E(\hat{s}_{ij}^{kt})$$

$$\eta_3 = (1.915, 2.307, 2.162)$$

**Step5**: Determine whether to perform group interaction, Determine whether the group consistency vector index satisfies the condition, and $\eta_{11} \succ 1.5, \eta_{12} \succ 2.5, \eta_{13} \succ 1.5$, hence we need to perform group interaction。

**Step6**: Calculate the entropy weight under each attribute according to the equation

Table 32 entropy weight under each attribute in the third round

| Supply chain | Attribute ($c_1$) | Attribute ($c_2$) | Attribute ($c_3$) | Attribute ($c_4$) |
|---|---|---|---|---|
| $a_1$ | $I_{11}^3=0.999$ | $I_{12}^3=0.997$ | $I_{13}^1=0.998$ | $I_{14}^1=0.995$ |
| $a_2$ | $I_{21}^3=0.999$ | $I_{22}^3=0.989$ | $I_{23}^1=0.975$ | $I_{24}^1=0.993$ |
| $a_3$ | $I_{31}^3=0.997$ | $I_{32}^3=0.997$ | $I_{33}^1=0.971$ | $I_{34}^1=0.998$ |
| $a_4$ | $I_{41}^3=0.987$ | $I_{42}^3=0.973$ | $I_{43}^3=0.972$ | $I_{44}^3=0.999$ |
| Average information entropy | $I_1^3=0.995$ | $I_2^3=0.989$ | $I_3^3=0.979$ | $I_4^3=0.996$ |
| Entropy weight | $w_1^3=0.251$ | $w_2^3=0.250$ | $w_3^3=0.247$ | $w_4^3=0.252$ |

Step 7: Adjust decision makers' preference information according to the equation, then obtain ta new round of decision makers' individual decision preference information, return to step 1.

**The Fourth round of group decision:**

Table 33 : The index values with expects four supply chain by experts e1 in the 4th round

| Supply chain | Attribute ($c_1$) | Attribute ($c_2$) | Attribute ($c_3$) | Attribute ($c_4$) |
|---|---|---|---|---|
| $a_1$ | $([\dot{s}_{4.808}, \dot{s}_{4.808}], [\ddot{s}_2, \ddot{s}_3])$ | $([\dot{s}_{2.371}, \dot{s}_{3.191}], [\ddot{s}_2, \ddot{s}_2])$ | $([\dot{s}_{3.812}, \dot{s}_{4.635}], [\ddot{s}_3, \ddot{s}_3])$ | $([\dot{s}_{3.603}, \dot{s}_{4.553}], [\ddot{s}_1, \ddot{s}_1])$ |
| $a_2$ | $([\dot{s}_{3.373}, \dot{s}_{4.192}], [\ddot{s}_2, \ddot{s}_3])$ | $([\dot{s}_{4.248}, \dot{s}_{4.619}], [\ddot{s}_2, \ddot{s}_2])$ | $([\dot{s}_{2.832}, \dot{s}_{3.197}], [\ddot{s}_3, \ddot{s}_3])$ | $([\dot{s}_{3.623}, \dot{s}_{3.634}], [\ddot{s}_1, \ddot{s}_1])$ |
| $a_3$ | $([\dot{s}_{2.373},\dot{s}_{3.373}],[\ddot{s}_2,\ddot{s}_3])$ | $([\dot{s}_{3.551},\dot{s}_{4.180}],[\ddot{s}_2,\ddot{s}_2])$ | $([\dot{s}_{2.457},\dot{s}_{3.280}],[\ddot{s}_3,\ddot{s}_3])$ | $([\dot{s}_{4.063},\dot{s}_{4.634}],[\ddot{s}_1,\ddot{s}_1])$ |
| $a_4$ | $([\dot{s}_{4.456},\dot{s}_{5.264}],[\ddot{s}_2,\ddot{s}_3])$ | $([\dot{s}_{1.752},\dot{s}_{2.752}],[\ddot{s}_2,\ddot{s}_2])$ | $([\dot{s}_{2.168},\dot{s}_{3.168}],[\ddot{s}_3,\ddot{s}_3])$ | $([\dot{s}_{3.453},\dot{s}_{4.178}],[\ddot{s}_1,\ddot{s}_1])$ |

Table 34: The index values with expects four supply chain by experts e2 in the 4th round

| Supply chain | Attribute $(c_1)$ | Attribute $(c_2)$ | Attribute $(c_3)$ | Attribute $(c_4)$ |
| --- | --- | --- | --- | --- |
| $a_1$ | $([\dot{s}_{4.394},\dot{s}_{4.394}],[\ddot{s}_2,\ddot{s}_3])$ | $([\dot{s}_{2.789},\dot{s}_{3.608}],[\ddot{s}_2,\ddot{s}_2])$ | $([\dot{s}_{3.385},\dot{s}_{4.207}],[\ddot{s}_3,\ddot{s}_3])$ | $([\dot{s}_{4.414},\dot{s}_{5.408}],[\ddot{s}_1,\ddot{s}_1])$ |
| $a_2$ | $([\dot{s}_{3.788},\dot{s}_{4.606}],[\ddot{s}_2,\ddot{s}_3])$ | $([\dot{s}_{2.995},\dot{s}_{3.783}],[\ddot{s}_2,\ddot{s}_2])$ | $([\dot{s}_{3.260},\dot{s}_{4.052}],[\ddot{s}_3,\ddot{s}_3])$ | $([\dot{s}_{2.609},\dot{s}_{3.206}],[\ddot{s}_1,\ddot{s}_1])$ |
| $a_3$ | $([\dot{s}_{2.788},\dot{s}_{3.788}],[\ddot{s}_2,\ddot{s}_3])$ | $([\dot{s}_{3.969},\dot{s}_{4.180}],[\ddot{s}_2,\ddot{s}_2])$ | $([\dot{s}_{2.030},\dot{s}_{2.852}],[\ddot{s}_3,\ddot{s}_3])$ | $([\dot{s}_{3.410},\dot{s}_{4.206}],[\ddot{s}_1,\ddot{s}_1])$ |
| $a_4$ | $([\dot{s}_{4.456},\dot{s}_{4.850}],[\ddot{s}_2,\ddot{s}_3])$ | $([\dot{s}_{3.005},\dot{s}_{4.005}],[\ddot{s}_2,\ddot{s}_2])$ | $([\dot{s}_{1.740},\dot{s}_{2.740}],[\ddot{s}_3,\ddot{s}_3])$ | $([\dot{s}_{3.802},\dot{s}_{4.178}],[\ddot{s}_1,\ddot{s}_1])$ |

Table 35: The index values with expects four supply chain by experts e3 in the 4th round

| Supply chain | Attribute $(c_1)$ | Attribute $(c_2)$ | Attribute $(c_3)$ | Attribute $(c_4)$ |
| --- | --- | --- | --- | --- |
| $a_1$ | $([\dot{s}_{4.808},\dot{s}_{4.808}],[\ddot{s}_2,\ddot{s}_3])$ | $([\dot{s}_{2.789},\dot{s}_{3.191}],[\ddot{s}_2,\ddot{s}_2])$ | $([\dot{s}_{3.812},\dot{s}_{4.207}],[\ddot{s}_3,\ddot{s}_3])$ | $([\dot{s}_{3.986},\dot{s}_{4.981}],[\ddot{s}_1,\ddot{s}_1])$ |
| $a_2$ | $([\dot{s}_{3.788},\dot{s}_{4.192}],[\ddot{s}_2,\ddot{s}_3])$ | $([\dot{s}_{3.830},\dot{s}_{4.619}],[\ddot{s}_2,\ddot{s}_2])$ | $([\dot{s}_{1.977},\dot{s}_{2.770}],[\ddot{s}_3,\ddot{s}_3])$ | $([\dot{s}_{3.037},\dot{s}_{3.206}],[\ddot{s}_1,\ddot{s}_1])$ |
| $a_3$ | $([\dot{s}_{2.788},\dot{s}_{3.788}],[\ddot{s}_2,\ddot{s}_3])$ | $([\dot{s}_{4.387},\dot{s}_{4.598}],[\ddot{s}_2,\ddot{s}_2])$ | $([\dot{s}_{1.602},\dot{s}_{1.997}],[\ddot{s}_3,\ddot{s}_3])$ | $([\dot{s}_{3.838},\dot{s}_{4.206}],[\ddot{s}_1,\ddot{s}_1])$ |
| $a_4$ | $([\dot{s}_{3.212},\dot{s}_{4.021}],[\ddot{s}_2,\ddot{s}_3])$ | $([\dot{s}_{2.170},\dot{s}_{3.170}],[\ddot{s}_2,\ddot{s}_2])$ | $([\dot{s}_{3.023},\dot{s}_{4.023}],[\ddot{s}_3,\ddot{s}_3])$ | $([\dot{s}_{3.802},\dot{s}_{4.606}],[\ddot{s}_1,\ddot{s}_1])$ |

**Step 1**: Based on the 2DULGA operator, we calculate the comprehensive value of each alternative, obtain the group decision preference matrix in the fourth round.

Table36   The index values with group decision makes in the forth round

| Attribute $(c_1)$ | Attribute $(c_2)$ | Attribute $(c_3)$ | Attribute $(c_4)$ |
| --- | --- | --- | --- |
| $([\dot{s}_{4.670},\dot{s}_{4.670}],[\ddot{s}_2,\ddot{s}_3])$ | $([\dot{s}_{2.649},\dot{s}_{3.330}],[\ddot{s}_2,\ddot{s}_2])$ | $([\dot{s}_{3.670},\dot{s}_{4.350}],[\ddot{s}_3,\ddot{s}_3])$ | $([\dot{s}_{4.001},\dot{s}_{4.981}],[\ddot{s}_1,\ddot{s}_1])$ |
| $([\dot{s}_{3.649},\dot{s}_{4.330}],[\ddot{s}_2,\ddot{s}_3])$ | $([\dot{s}_{3.691},\dot{s}_{4.340}],[\ddot{s}_2,\ddot{s}_2])$ | $([\dot{s}_{2.690},\dot{s}_{3.340}],[\ddot{s}_3,\ddot{s}_3])$ | $([\dot{s}_{3.090},\dot{s}_{3.349}],[\ddot{s}_1,\ddot{s}_1])$ |
| $([\dot{s}_{2.649},\dot{s}_{3.649}],[\ddot{s}_2,\ddot{s}_3])$ | $([\dot{s}_{3.969},\dot{s}_{4.320}],[\ddot{s}_2,\ddot{s}_2])$ | $([\dot{s}_{2.030},\dot{s}_{2.710}],[\ddot{s}_3,\ddot{s}_3])$ | $([\dot{s}_{3.770},\dot{s}_{4.349}],[\ddot{s}_1,\ddot{s}_1])$ |
| $([\dot{s}_{4.041},\dot{s}_{4.712}],[\ddot{s}_2,\ddot{s}_3])$ | $([\dot{s}_{2.309},\dot{s}_{3.309}],[\ddot{s}_2,\ddot{s}_2])$ | $([\dot{s}_{2.310},\dot{s}_{3.310}],[\ddot{s}_3,\ddot{s}_3])$ | $([\dot{s}_{3.686},\dot{s}_{4.321}],[\ddot{s}_1,\ddot{s}_1])$ |

**Step2: calculate the distance matrix between the** individual experts matrix and group decision in the fourth round

Table37   **the distance matrix between the** individual experts e1 and group decision in the 4-th round

| Supply chain | Attribute $(c_1)$ | Attribute $(c_2)$ | Attribute $(c_3)$ | Attribute $(c_4)$ |
| --- | --- | --- | --- | --- |
| $a_1$ | $\hat{d}_{11}^{14}$ =0.014 | $\hat{d}_{12}^{14}$ =0.017 | $\hat{d}_{13}^{14}$ =0.027 | $\hat{d}_{14}^{14}$ =0.017 |
| $a_2$ | $\hat{d}_{21}^{14}$ =0.022 | $\hat{d}_{22}^{14}$ =0.035 | $\hat{d}_{23}^{14}$ =0.018 | $\hat{d}_{24}^{14}$ =0.017 |
| $a_3$ | $\hat{d}_{31}^{14}$ =0.029 | $\hat{d}_{32}^{14}$ =0.023 | $\hat{d}_{33}^{14}$ =0.062 | $\hat{d}_{34}^{14}$ =0.012 |
| $a_4$ | $\hat{d}_{41}^{14}$ =0.050 | $\hat{d}_{42}^{14}$ =0.046 | $\hat{d}_{43}^{14}$ =0.018 | $\hat{d}_{44}^{14}$ =0.008 |

Table 38 **the distance matrix between the** individual experts e2 and group decision in the 4-th round

| Supply chain | Attribute $(c_1)$ | Attribute $(c_2)$ | Attribute $(c_3)$ | Attribute $(c_4)$ |
|---|---|---|---|---|
| $a_1$ | $\hat{d}_{11}^{24}$ =0.029 | $\hat{d}_{12}^{24}$ =0.017 | $\hat{d}_{13}^{24}$ =0.027 | $\hat{d}_{14}^{24}$ =0.059 |
| $a_2$ | $\hat{d}_{21}^{24}$ =0.022 | $\hat{d}_{22}^{24}$ =0.052 | $\hat{d}_{23}^{24}$ =0.080 | $\hat{d}_{24}^{24}$ =0.035 |
| $a_3$ | $\hat{d}_{31}^{24}$ =0.014 | $\hat{d}_{32}^{24}$ =0.006 | $\hat{d}_{33}^{24}$ =0.009 | $\hat{d}_{34}^{24}$ =0.042 |
| $a_4$ | $\hat{d}_{41}^{24}$ =0.029 | $\hat{d}_{42}^{24}$ =0.058 | $\hat{d}_{43}^{24}$ =0.071 | $\hat{d}_{44}^{24}$ =0.044 |

Table 39 **the distance matrix between the** individual experts e3 and group decision in the 4-th round.

| Supply chain | Attribute $(c_1)$ | Attribute $(c_2)$ | Attribute $(c_3)$ | Attribute $(c_4)$ |
|---|---|---|---|---|
| $a_1$ | $\hat{d}_{11}^{34}$ =0.014 | $\hat{d}_{12}^{34}$ =0.012 | $\hat{d}_{13}^{34}$ =0.018 | $\hat{d}_{14}^{34}$ =0.000 |
| $a_2$ | $\hat{d}_{21}^{34}$ =0.014 | $\hat{d}_{22}^{34}$ =0.017 | $\hat{d}_{23}^{34}$ =0.080 | $\hat{d}_{24}^{34}$ =0.004 |
| $a_3$ | $\hat{d}_{31}^{34}$ =0.014 | $\hat{d}_{32}^{34}$ =0.029 | $\hat{d}_{33}^{34}$ =0.071 | $\hat{d}_{34}^{34}$ =0.004 |
| $a_4$ | $\hat{d}_{41}^{34}$ =0.079 | $\hat{d}_{42}^{34}$ =0.012 | $\hat{d}_{43}^{34}$ =0.089 | $\hat{d}_{44}^{34}$ =0.008 |

Step3: calculate the expectation of two-dimensional language evaluation information $E(\hat{s}_{ij}^{kt})$ according to the follow equation。

$$E(\hat{s}_{ij}^{kt}) = \frac{a_{ij}^{kt} + b_{ij}^{kt}}{2*(l-1)} * \frac{c_{ij}^{kt} + d_{ij}^{kt}}{2*(z-1)} \ (i, j = 1,2,3,4; k = 1,2,3)$$

Table 40   the expectation value of expert e1 in the 4-th round

| Supply chain | Attribute $(c_1)$ | Attribute $(c_2)$ | Attribute $(c_3)$ | Attribute $(c_4)$ |
|---|---|---|---|---|
| $a_1$ | $E(\hat{s}_{11}^{14})$=0.501 | $E(\hat{s}_{12}^{14})$=0.232 | $E(\hat{s}_{13}^{14})$=0.528 | $E(\hat{s}_{14}^{14})$=0.170 |
| $a_2$ | $E(\hat{s}_{21}^{14})$=0.394 | $E(\hat{s}_{22}^{14})$=0.369 | $E(\hat{s}_{23}^{14})$=0.377 | $E(\hat{s}_{24}^{14})$=0.151 |
| $a_3$ | $E(\hat{s}_{31}^{14})$=0.299 | $E(\hat{s}_{32}^{14})$=0.322 | $E(\hat{s}_{33}^{14})$=0.359 | $E(\hat{s}_{34}^{14})$=0.181 |
| $a_4$ | $E(\hat{s}_{41}^{14})$=0.506 | $E(\hat{s}_{42}^{14})$=0.188 | $E(\hat{s}_{43}^{14})$=0.333 | $E(\hat{s}_{44}^{14})$=0.159 |

Table 41 the expectation value of expert e2 in the 4-th round

| Supply chain | Attribute $(c_1)$ | Attribute $(c_2)$ | Attribute $(c_3)$ | Attribute $(c_4)$ |
|---|---|---|---|---|
| $a_1$ | $E(\hat{s}_{11}^{24})=0.458$ | $E(\hat{s}_{12}^{24})=0.267$ | $E(\hat{s}_{13}^{24})=0.475$ | $E(\hat{s}_{14}^{24})=0.205$ |
| $a_2$ | $E(\hat{s}_{21}^{24})=0.437$ | $E(\hat{s}_{22}^{24})=0.282$ | $E(\hat{s}_{23}^{24})=0.457$ | $E(\hat{s}_{24}^{24})=0.121$ |
| $a_3$ | $E(\hat{s}_{31}^{24})=0.342$ | $E(\hat{s}_{32}^{24})=0.340$ | $E(\hat{s}_{33}^{24})=0.305$ | $E(\hat{s}_{34}^{24})=0.159$ |
| $a_4$ | $E(\hat{s}_{41}^{24})=0.485$ | $E(\hat{s}_{42}^{24})=0.292$ | $E(\hat{s}_{43}^{24})=0.280$ | $E(\hat{s}_{44}^{24})=0.166$ |

Table 42 the expectation value of expert e3 in the 4-th round

| Supply chain | Attribute $(c_1)$ | Attribute $(c_2)$ | Attribute $(c_3)$ | Attribute $(c_4)$ |
|---|---|---|---|---|
| $a_1$ | $E(\hat{s}_{11}^{34})=0.501$ | $E(\hat{s}_{12}^{34})=0.249$ | $E(\hat{s}_{13}^{34})=0.501$ | $E(\hat{s}_{14}^{34})=0.187$ |
| $a_2$ | $E(\hat{s}_{21}^{34})=0.416$ | $E(\hat{s}_{22}^{34})=0.352$ | $E(\hat{s}_{23}^{34})=0.297$ | $E(\hat{s}_{24}^{34})=0.130$ |
| $a_3$ | $E(\hat{s}_{31}^{34})=0.342$ | $E(\hat{s}_{32}^{34})=0.374$ | $E(\hat{s}_{33}^{34})=0.225$ | $E(\hat{s}_{34}^{34})=0.168$ |
| $a_4$ | $E(\hat{s}_{41}^{34})=0.377$ | $E(\hat{s}_{42}^{34})=0.223$ | $E(\hat{s}_{43}^{34})=0.440$ | $E(\hat{s}_{44}^{34})=0.175$ |

**Step 4**: calculate the group consistency index according to the following equation to obtain the vector of group consistency index

$$\eta_k^t = \sum_{i=1}^{m}\sum_{j=1}^{n} d(\hat{s}_{ij}^{kt}, \hat{y}_{ij}^{kt}) / E(\hat{s}_{ij}^{kt})$$

$$\eta_4 = (1.422, 2.304, 1.472)$$

**Step5**: Determine whether to perform group interaction, Determine whether the group consistency vector index satisfies the condition.

Table 43 Group consistency index values of each round group decision

|  | Group consistency index e1 | Group consistency index e1 | Group consistency index e3 |
|---|---|---|---|
| 1st round | $\eta_1^1=4.464$ | $\eta_2^1=6.776$ | $\eta_3^1=4.408$ |
| 2nd round | $\eta_1^2=2.583$ | $\eta_2^2=3.382$ | $\eta_3^2=2.294$ |
| 3rd round | $\eta_1^3=1.915$ | $\eta_2^3=3.307$ | $\eta_3^3=2.162$ |
| 4th round | $\eta_1^4=1.422$ | $\eta_2^4=2.304$ | $\eta_3^4=1.472$ |

Through the change of $\{\eta_t\}$, it is found that the index value $\{\eta_t\}$ is gradually reduced, which means that the dynamic group decision making method can ensure the effectiveness of the group interaction process and achieve the consensus evaluation of the group.